\documentclass[doc,a4paper]{article}

\usepackage{graphicx} 
\usepackage{palatino}
\usepackage{fancyhdr}
\usepackage{gensymb}

\usepackage[english]{babel}
\usepackage{amsmath,bbm}
\usepackage{amsfonts}
\usepackage{amssymb}
\usepackage{graphicx}
\usepackage{color}
\usepackage{url}
\usepackage{algorithm}
\usepackage{tikz,pgfplots}
\usetikzlibrary{shapes,shapes.multipart,shapes.geometric}
\usetikzlibrary{matrix,shapes,arrows,positioning,chains,calc,decorations.pathreplacing}
\tikzstyle{line} = [draw, -latex']

\usepackage{algpseudocode}
\usepackage{letltxmacro}
\LetLtxMacro{\oldalgorithmic}{\algorithmic}
\LetLtxMacro{\endoldalgorithmic}{\endalgorithmic}
\renewenvironment{algorithmic}[1][0]{%
  \hrulefill\par
  \oldalgorithmic[#1]}
  {\endoldalgorithmic\par
   \vspace*{-.5\baselineskip}
   \hrulefill\par
  }
  
\newtheorem{remark}{Remark}

\newcommand{\R}{\mathbbm R}

\newcommand{\abs}[1]{\left| #1 \right|}

\title{Dynamical optical flow of saliency maps for predicting visual attention}

\begin{document}
\author{
  Aniello Raffaele Patrone\\
Computational Science Center, University of Vienna\\
  \and
  Christian Valuch\\
  Faculty of Psychology, University of Vienna\\
  Faculty of Biology and Psychology, University of G\"ottingen
  \and
  Ulrich Ansorge\\
  Faculty of Psychology, University of Vienna
  \and
  Otmar Scherzer\\
  Computational Science Center, University of Vienna\\
  Johann Radon Institute for Computational \\
  and Applied Mathematics (RICAM)
}
\maketitle

\begin{abstract}

Saliency maps are used to understand human attention and visual fixation. However, while very well established 
for static images, there is no general agreement on how to compute a saliency map of dynamic scenes. 
In this paper we propose a mathematically rigorous approach to this problem, 
including static saliency maps of each video frame for the calculation of the optical flow. 
Taking into account static saliency maps for calculating the optical flow allows for overcoming the aperture problem. 
Our approach is able to explain human fixation behavior in situations which pose challenges to standard approaches, such as when a fixated object disappears behind an occlusion and reappears after several frames. In addition, we quantitatively compare our model against alternative solutions using a large eye tracking data set. Together, our results suggest that assessing optical flow information across a series of saliency maps gives a highly accurate and useful account of human overt attention in dynamic scenes.
\end{abstract}

%

\lhead{
%
}
\thispagestyle{plain}

\section{Introduction} 

Humans and other primates focus their perceptual and cognitive processing on aspects of the visual input. 
This selectivity is known as \textit{visual attention} and it is closely linked to eye movements: 
humans rapidly shift their center of gaze multiple times per second from one location of an image to another. These gaze 
shifts are called saccades, and they are necessary because high acuity vision is limited to a small central area of the visual 
field. Fixations are the periods between two saccades, in which the eyes rest relatively still on one location from which 
information is perceived with high acuity. Consequently, fixations reflect which areas of an image attract the viewer's attention.

Models of visual attention and eye behavior can be roughly categorized into two classes: bottom-up models, which are 
task-independent and 
driven mainly by the intrinsic features of the visual stimuli, and top-down models, which are task-dependent and driven 
by high-level 
processes \cite{Hen03}. Our focus in this article is on bottom-up models.
A central concept in bottom-up models of attention is the saliency map, which is a topographical 
representation of the original image representing the probability of each location to attract the 
viewer's attention. 
Saliency maps are useful for testing hypotheses on the importance of the current image's low-level visual features (such as color, luminance, or orientation). 
Moreover, saliency models allow for general predictions on the locations that are fixated by human viewers. 
This is important in many contexts, such as the optimization of video compression algorithms or of graphical user interfaces at the workplace as well 
as in entertainment environments, to name but a few examples.
A location in an image is considered \emph{salient} if it stands out compared to its local surroundings. 

Saliency maps are often computed based on low-level visual features and their local contrast strengths 
in dimensions such as color, luminance, or orientation. A well-known model for \emph{static} scenes is the one 
of \cite{IttKocNie98}. Other examples are graph-based visual saliency (GBVS)\cite{HarKocPer06}, 
gaze-attentive fixation finding engine (GAFFE)\cite{RajVanBov08}, frequency-tuned saliency detection model 
\cite{AchHemEstSus09} and models based on phase spectrum, explained by the inverse Fourier transform \cite{GuoZha10}.

Recent work is devoted to develop concepts of saliency maps of \emph{dynamic} sequences
\cite{FanWanLinFan14,IttBal05b,IttBal05a,IttKoc01,LiuYueQiu09,RenGaoChiRaj13}. 
These \emph{spatial-temporal saliency maps} are modeled as the weighted sum of  motion features and of static saliency maps
\cite{FanWanLinFan14,IttBal05a,IttBal05b,IttKoc01,LiLee07,LiuYueQiu09,MaZha02,MarHoGraGuyPelGue08,PetItt07,RenGaoChiRaj13,SchuBraGeg09,ZhaTonCot09}.

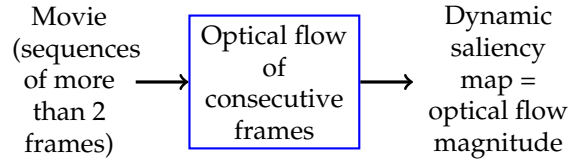
\begin{figure}[t]
\begin{center}
\begin{tikzpicture}
\matrix [column sep=7mm, row sep=5mm] {
  \node (in)  {\parbox{1.5cm}{\centering Movie\\ (sequences of more than 2 frames)}}; &
          \node (of) [draw=blue,thick, shape=rectangle] {\parbox{2cm}{\centering Optical flow of consecutive frames}}; &
  \node (out) {\parbox{2cm}{\centering Dynamic saliency map =\\ optical flow magnitude}};\\         
};
\draw[->,very thick] (in) -- (of);
\draw[->,very thick] (of) -- (out);
\end{tikzpicture}
\caption{The standard approach for calculating a dynamic saliency map}
\label{fig:algSt}
\end{center}
\end{figure}

The present work introduces a \emph{novel} dynamic saliency map, which is the optical flow of a high-dimensional dynamic sequence. Extending the concept of saliency to dynamic sequences by including optical flow as an 
additional source for bottom-up saliency is not new per se 
\cite{gao2004discriminant,MarHoGraGuyPelGue08,tsotsos1995modeling,vijayakumar2001overt}. 
However, while other researchers use the optical flow as a feature of the dynamic saliency map, we define the dynamic 
saliency map as the optical flow itself. In detail:
\begin{enumerate}
 \item we calculate the flow of a \emph{virtual}, \emph{high-dimensional} image sequence,  
which consists of (i) intensity and (ii) color channels, \emph{complemented} by saliency maps, respectively;
\item we also consider the complete movie (consisting of all frames) for the computation of a dynamic saliency map.
In contrast, in \cite{FanWanLinFan14,IttBal05a,IttBal05b,IttKoc01,LiLee07,LiuYueQiu09,MaZha02,MarHoGraGuyPelGue08,PetItt07,RenGaoChiRaj13,SchuBraGeg09,
ZhaTonCot09} (as it is standard), dynamic saliency maps are obtained from 
optical flow features (see Figure \ref{fig:algSt}) of \emph{two consecutive} frames. 
As we show below in section \ref{sec:results} this can lead to misinterpretations of visual attention, for instance, in the case of occlusions.
\end{enumerate}

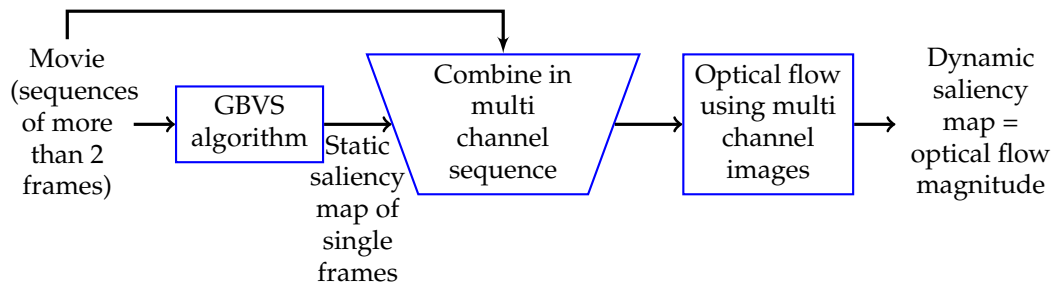
\begin{figure}[b]
\hspace{-1.2cm}
\begin{tikzpicture}
\matrix [column sep=5.5mm, row sep=5mm] {
  \node (in)  {\parbox{1.5cm}{\centering Movie\\ (sequences of more than 2 frames)}}; &
  \node (gb) [draw=blue,thick, shape=rectangle] {\parbox{1.7cm}{\centering GBVS\\ algorithm}}; &
  \node (mc) [draw=blue,thick, shape=trapezium, trapezium left angle=110,
        trapezium right angle=110] {\parbox{2cm}{\centering Combine in multi\\channel sequence}}; &
          \node (of) [draw=blue,thick, shape=rectangle] {\parbox{2cm}{\centering Optical flow using multi\\channel images}}; &
  \node (out) {\parbox{2cm}{\centering Dynamic saliency map =\\ optical flow magnitude}};\\         
};
\draw[->,very thick] (in) -- (gb);
\draw[->,very thick] (gb) -- (mc) node[pos=0.5,below]{\parbox{2cm}{\centering Static saliency map of single frames}};
\path [line,very thick] (in)-- ++ (0,+1.5) -| (mc) ;
\draw[->,very thick] (mc) -- (of);
\draw[->,very thick] (of) -- (out);
\end{tikzpicture}
\caption{The proposed approach for calculating a dynamic saliency map.}
\label{fig:algorithm}
\end{figure}

Our algorithm for calculating the dynamic saliency map is schematically depicted in Figure \ref{fig:algorithm}.
From the different methods proposed in the literature to estimate optical flow, we focus on variational methods, which are 
key methods in computational image processing and computer vision.

The outline of this paper is as follows: in section \ref{sec:compmeth} we review the optical flow, 
introduce the new model and derive a fixed-point algorithm for the computational realization;
in section \ref{sec:eyeTrack} we discuss the acquisition of the eye tracking data;
finally, in sections \ref{sec:results} and \ref{sec:concl}
we present experiments, results and a discussion.

\section{Computational methods}
\label{sec:compmeth}

The optical flow denotes the pattern of apparent motion of objects and surfaces in a dynamic sequence.
The computational model of optical flow is based on the \emph{brightness-constancy assumption}, requiring that for 
every pixel there exists a path through the movie which conserves brightness.

\subsection{Basic optical flow calculations}
We briefly outline the concept in a continuous mathematical formulation.  
We consider a \emph{movie} to be a time continuous recording of images, where each image is described by a function defined 
for $x=(x_1,x_2)^t$ in the Euclidean plane $\R^2$. This function representing the image is called \emph{frame}. 
Moreover, we assume that the movie to be analyzed has unit-length in time. That is, the movie can be parametrized 
by a time $t\in[0,1]$. If the frames composing a movie consist of gray-valued images, then we describe each by a function 
$f:\R^2 \to \R$. 
If the frame is completed by a spatial saliency map, then $\vec{f} := (f_1;f_2)^t:\R^2 \to \R^2$, where $f_1$ is the 
recorded movie and $f_2$ is the according saliency map. 
For a color frame, $\vec{f} := (f_1,f_2,f_3)^t:\R^2 \to \R^3$, where each component represents a channel of the 
color images; typically RGB (red-green-blue) or HSV (hue-saturation-value) channels. 
A color frame, which is complemented by a saliency map is described by 
$\vec{f} := (f_1,f_2,f_3;f_4)^t:\R^2 \to \R^4$, where the first three components are the color channels and the 
fourth component is the according saliency map. For the sake of simplicity of notation, from now on, we will always 
write $\vec{f}$, even if $f$ is a gray-valued image.

The optical flow equation is derived from the \emph{brightness constancy assumption}, which considers paths $\gamma$ 
of constant intensity of the movie or the saliency complemented movie, respectively. Note that in our setting we consider 
brightness in each component. That is for 
\begin{equation}
\label{eq:bca}
\vec{f}(\gamma(x,t),t)=\vec{c}
\end{equation}
for some constant vector $\vec{c}$. Note that we do not differentiate in our notation between intensity, color, or saliency 
completed movies anymore and always write $\vec{f}$ for the image data.

A differential formulation of the brightness constancy assumption follows 
from \eqref{eq:bca} by differentiation with respect to time (see for instance \cite{HorSchu81}):
\begin{equation}
\label{eq:ofe}
J_{\vec{f}\,}(x,t) \cdot \vec{u}(x,t) + \frac{d \vec{f}}{d t}(x,t)=0 \text{ for all } x \in \R^2 \text{ and } t \in (0,1)\,,
\end{equation}
where $\vec{u}(x,t)=(u_1(x,t),u_2(x,t))=\frac{d \gamma}{d t}(x,t)$ is the optical flow and 
$J_{\vec{f}\,},\frac{d \vec{f}}{d t}$ are the 
partial derivatives in space and time of the function $\vec{f}$, respectively. Note that $(u_1(x,t),u_2(x,t))$ can both be 
vectors, and that the \emph{Jacobian} $J_{\vec{f}\,} = \nabla \vec{f} = 
\left(\frac{\partial}{\partial x_1} \vec{f}, \frac{\partial}{\partial x_2} \vec{f}\, \right)$ is a two-dimensional vector if the 
movie is gray-valued, a $(2 \times 2)$-dimensional matrix if it is a saliency complemented gray-valued image,
a $(3 \times 2)$-dimensional matrix if it is in color, and a $(4 \times 2)$-dimensional matrix if a color image 
is complemented by a saliency map.

Equation \eqref{eq:ofe} is uniquely solvable at points $(x,t)$ where $J_{\vec{f}\,}$ has full-rank $2$. 
For gray-valued movies the matrix can have at most rank of one, and thus the two unknown functions $u_1$ and $u_2$ \emph{cannot}
be reconstructed uniquely from this equation. This is known as the \emph{aperture problem} in Computer Vision. 
The non-uniqueness is taken care of mathematically by restricting attention to the \emph{minimum energy solution} 
of \eqref{eq:ofe} which minimizes, among all solutions, an appropriately chosen 
energy, such as the one proposed in \cite{HorSchu81}:
\begin{equation}
\label{eq:hsr}
\mathcal{E}[t](\vec{u}):=\int_{\R^2} |\nabla u_1(x,t)|^2+|\nabla u_2(x,t)|^2 dx \text{ for all } t \in [0,1].
\end{equation}
For every $t \in (0,1)$ the minimum energy solution can be approximated (see for instance 
\cite{SchGraGroHalLen09} for a rigorous mathematical statement) by the minimizer of
\begin{equation}
\label{eq:hsm_o}
\begin{aligned}
 \mathcal{F}[t](\vec{u}):=&\int_{\R^2} \abs{J_{\vec{f}\,}(x,t)\cdot \vec{u}(x,t)+\frac{d \vec{f}}{d t}(x,t)}^2 dx\\
&+ \alpha \int_{\R^2} |\nabla u_1(x,t)|^2+|\nabla u_2(x,t)|^2 dx.
\end{aligned}
\end{equation}
Here $\alpha > 0$ is a weight (also called regularization) parameter. 

Different optical flow methods have been considered in the literature, which are formulated via different regularization energies.
For instance, in \cite{WeiSchn01b} it was suggested to minimize
\begin{equation}
\label{eq:wsST}
\begin{aligned}
\int_{\R^2} \abs{J_{\vec{f}\,}(x,t) \cdot \vec{u}(x,t)+\frac{d \vec{f}}{d t}(x,t)}^2 
&+ \alpha \Psi(|\nabla u_1(x,t)|^2+|\nabla u_2(x,t)|^2) dx\\ 
&\text{ for all } t \in [0,1],
\end{aligned}
\end{equation}
where $\Psi: \R_0^+ \rightarrow \R_0^+$ is a monotonically increasing, differentiable function. 
Note that in the original work of \cite{WeiSchn01b} the function $\vec{f}$ is not saliency complemented and the data 
$\vec{f}$ only represents intensity images. 

We note that in most applications, $\mathcal{F}[t]$ is realized by replacing $J_{\vec{f}\,}$ by the 
spatial difference quotients, $\frac{d \vec{f}}{d t}$ by the temporal difference of 
two consecutive frames $\vec{f}{(x,t_i)}$, $i=1,2$, respectively, and appropriate scaling. 
This results in the functional to be minimized:
\begin{equation}
\label{eq:hsm}
\begin{aligned}
 &\int_{\R^2} \abs{J_{{\vec{f}(\cdot,t_1)\,}}(x)\cdot \vec{u}(x)+\vec{f}(x,t_2) - \vec{f}(x,t_1)}^2 dx \\
&+ \alpha \int_{\R^2} \Psi(|\nabla u_1(x)|^2+|\nabla u_2(x)|^2)dx.
\end{aligned}
\end{equation}
\begin{remark}
\label{re:coarse}
We emphasize that by the approximation of $\frac{d \vec{f}}{d t}$ with the finite difference 
$\vec{f}(x,t_2) - \vec{f}(x,t_1)$ (where, we assume that after time scaling 
we have $t_2-t_1 = 1$), the equations \eqref{eq:ofe} and 
\begin{equation}
\label{eq:ofeII}
J_{\vec{f}(\cdot,t_1)\,}(x) \cdot \vec{u}(x) + \vec{f}(x,t_2) - \vec{f}(x,t_1)=0 \text{ for all } x \in \R^2\,,
\end{equation}
might not be correlated anymore. This is especially true for large displacements $\gamma - I$. 
In order to overcome this discrepancy, researchers in computer vision  proposed to use computational coarse-to-fine strategies \cite{Ana89,BlaAnd96,BruWeiSchn05,MemPer98a,MemPer98b,MemPer02}. 
\end{remark}

We emphasize that in all these approaches the temporal coherence of the movie is neglected and this can lead to rather abruptly
changing flow sequences. Therefore, in \cite{WeiSchn01b}, a \emph{spatial-temporal} regularization was suggested, which 
consists in minimization of 
\begin{equation}
\label{eq:wsCT}
\begin{aligned}
\int_{\R^2 \times[0,1]} \abs{J_{\vec{f}\,} (x,t) \cdot \vec{u}(x,t)+\frac{d \vec{f}}{d t}(x,t)}^2
+\alpha  \Psi(|\nabla_3 u_1(x,t)|^2+|\nabla_3 u_2(x,t)|^2) dx dt,
\end{aligned}
\end{equation}
where $\nabla_3=(\nabla, \frac{\partial}{\partial t})$ denotes the spatial-temporal gradient operator. 
More sophisticated spatial-temporal regularization approaches have been proposed in 
\cite{AndSchZul15,BorItoKun03,PatSch15,WanFanWan08,WeiSchn01a,WeiSchn01b}.

\subsection{Spatial saliency for optical flow computations}
In the following section we investigate the effect of complementing intensity and color images with 
a spatial saliency map on optical flow computations. We use the GBVS method, introduced by \cite{HarKocPer06}, which 
defines a spatial saliency for each frame. Note however, that our approach is not restricted to this particular choice.

We verify the hypothesis that the complemented data uniquely determines the optical flow in 
selected regions - that is, where $J_{\vec{f}}$ has full rank. A conceptually similar strategy was implemented in 
\cite{MarHoGraGuyPelGue08} where the image data was complemented by Gabor filters as input of \eqref{eq:ofe}.
Here, we verify this thesis by checking the spatially dependent condition number of $J_{\vec{f}}$ in the plane, 
and by tabbing the area of the points in a region where the condition number is below 1000 (see Table \ref{tab:compCandG}).
In these regions we expect that the flow can be computed accurately from \eqref{eq:ofe} without any regularization.
We recall that for gray-valued images the optical flow equation is under-determined and the matrix $J_{\vec{f}}$ is 
singular, or in other words, the condition number is infinite. 
However, if the intensity data is complemented by a saliency map, $3 \text{ to }17\%$ of the pixels have a condition number 
smaller than 1000, such that the solution of \eqref{eq:ofe} can be determined 
in a numerically stable way. For color images the optical flow equation is already over-determined 
and $1\text{ to }5\%$ of the pixels have a small condition number.
However, if the color information is complemented by a saliency information, $3\text{ to }33\%$ of the pixels have this 
feature.
These results suggest that complementing the original information by saliency is useful for accurate 
computations of the optical flow. 

\begin{table}[tb]
\begin{center}
 \begin{tabular}{lccc}\hline
Sequence  & Saliency-Brightness  & Saliency-Color & Color   \\ \hline
 Backyard     & 3.16\%    & 8.18\%  & 2.45\%   \\
 Basketball     & 3.68\%    & 10.06\%  & 1.26\%   \\  
 Beanbags     & 4.07\%    & 9.73\%  & 0.9\%   \\
 Dimetrodon      &    5.91\%      &    7.54\%    &   1.07\%   \\ 
 DogDance    & 3.99\%   & 9.57\%  & 2.50\%   \\
 Dumptruck    & 5.20\%   & 10.13\%  & 3.35\%   \\
 Evergreen    & 6.12\%   & 12.48\%  & 3.20\%   \\ 
 Grove      &    12.61\%     &    18.94\%  &    4.86\%      \\ 
 Grove2      &    15.48\%     &    24.07\%  &    3.29\%      \\ 
 Grove3      &    16.49\%      &    25.27\%  &    4.20\%    \\ 
 Hydrangea      &    10.30\%      &   29.93\%  &   2.48\%       \\ 
 Mequon      &    11.44\%       &    19.12\%  &   2.38\%     \\ 
 MiniCooper      &   7.35\%       &    16.77\%  &   3.76\%     \\ 
 RubberWhale      &    8.45\%      &    22.34\%   &   1.46\%    \\ 
 Schefflera      &    12.12\%      &    20.02\%   &   1.98\%    \\ 
 Teddy      &    16.65\%       &   32.96\%   &   4.63\%   \\ 
 Urban2      &    3.17\%       &   3.35\%   &   0.31\%   \\ 
 Venus      &    12.86\%      &   17.84\%  &   2.75\%       \\ 
 Walking      &    2.85\%      &   7.41\%   &   1.25\%     \\  
 Wooden      &    3.38\%      &   7.76\%   &   0.75\%     \\ 
\hline     
 \end{tabular}
  \caption{Percentage of the pixels with condition number smaller than 1000 for different sequences from the Middleburry Dataset \protect\cite{BakSchRotBlaSze11}. For each sequence we use the third frame and its spatial saliency as input.}
   \label{tab:compCandG}
\end{center}
\end{table}

\subsection{Contrast invariance}
We also aim to recognize motion under varying illumination conditions because
humans have this visual capacity. 
However, a typical optical flow model, like the one in \eqref{eq:wsCT}, would not yield contrast invariant optical flow although this would be necessary for motion recognition under varying illumination conditions.

In order to restore contrast invariance for the minimizer of the optical flow functional, as proposed (for a different reason) by \cite{IglKir15} and by  \cite{LaiVem98, ZimBruWei11}, 
we introduce
the semi-norm $\|\cdot\|_\mathcal{B}^2$ like: $$\|w\|_\mathcal{B}^2=w^T \text{diag}(b_1,...,b_n) w\,,$$
where $b_i$ are the components of a vector $\mathcal{B}$. 
For gray-valued images complemented with saliency $\vec{f}=(f_{1};f_{2})$ the vector $\mathcal{B}$ is defined as:
\begin{equation}
\label{eq:cigs}
\mathcal{B}=  
\begin{pmatrix}
\frac{f_{2}}{\sqrt{|\nabla f_{1}|^2+\xi^2}} ,1
\end{pmatrix}.
\end{equation}
For saliency complemented color images $\vec{f}=(f_{1},f_{2},f_{3};f_{4})$ we define
\begin{equation}
\label{eq:cics}
\mathcal{B}=\begin{pmatrix}
\frac{f_{4}}{\sqrt{|\nabla f_{1}|^2}},
\frac{f_{4}}{\sqrt{|\nabla f_{2}|^2+\xi^2}},
\frac{f_{4}}{\sqrt{|\nabla f_{3}|^2+\xi^2}},
1
\end{pmatrix}.
\end{equation}
The vectors $\mathcal{B}$ are the product of the weighting factors $\frac{1}{\sqrt{|\nabla f_{i}|^2+\xi^2}}$ and the 
saliency map, respectively. This  means that features with high spatial saliency will be weighted stronger, and thus more emphasis on a precise optical flow calculation is given to these regions. 
We are then using the weighted semi-norm as an error measure of the residual of \eqref{eq:ofe}:
\begin{equation}
\label{eq:omSBF}
\int_{\R^2 \times[0,1]} \left\|J_{\vec{f\,}} (x,t) \cdot \vec{u}(x,t)+\frac{d \vec{f}}{d t}(x,t)\right \|_\mathcal{B}^2\;.
\end{equation} 
Finally, one needs to choose the constant $\xi$ for the denominator of each weighting factor. 
As in \cite{IglKir15} we choose  $\xi=0.01$.

\subsection{The final model}
We use as a regularization functional
\begin{equation*}
 \int_{\R^2 \times [0,1]} \Psi(|\nabla_3 u_1(x,t)|^2+|\nabla_3 u_2(x,t)|^2) dx dt
\end{equation*}
as in \cite{WeiSchn01b}, with the difference that the function  
$\Psi(r^2)=\epsilon r^2 + (1-\epsilon)\lambda^2 \sqrt{1+\frac{r^2}{\lambda^2}}$ is replaced by 
\begin{equation}
\label{eq:psiR}
\Psi(r^2)=\sqrt{r^2+\epsilon^2} \text{ with } \epsilon=10^{-6}
\end{equation}
as in \cite{BroBruPapWeo04}.
Moreover, we substitute the spatial-temporal gradient operator in \cite{WeiSchn01b} with $\nabla_3=(\nabla,\lambda \frac{\partial}{\partial t})$. 
In numerical realizations, the weighting parameter $\lambda$ corresponds to
the ratio of the sampling in space squared and the one in time
The resulting model for optical flow computations then consists in minimization of the functional
\begin{equation}
\label{eq:om}
\begin{aligned}
\int_{\R^2 \times[0,1]} \left \|{J_f (x,t) \cdot \vec{u}(x,t)+\frac{d \vec{f}}{d t}(x,t)} \right\|_{\mathcal{B}}^2
+\alpha  \Psi(|\nabla_3 u_1(x,t)|^2+|\nabla_3 u_2(x,t)|^2) dx dt.
\end{aligned}
\end{equation}
.

\subsubsection{Numerical solution}
The ultimate goal is to find the path $\gamma$ solving \eqref{eq:bca}, that is connecting a movie sequence. 
By applying Taylor expansion one sees that $\vec{u} \sim \frac{d \gamma}{d t}(x,t) t$ for small $t$. As we have stated in 
Remark \eqref{re:coarse}, when trying to find approximations of $\gamma$ via minimization of the proposed functional 
\eqref{eq:om} a coarse-to-fine strategy \cite{Ana89,BlaAnd96,BruWeiSchn05,MemPer98a,MemPer98b,MemPer02} is useful for the following two reasons:

First, since at a coarse level of discretization large displacements appear relatively small, the optical flow equation 
\eqref{eq:ofe}, which is a 
linearization of the brightness constancy equation \eqref{eq:bca}, is a good approximation of that. Indeed, a displacement 
of one pixel at the coarsest level of a 4-layer pyramid can represent 4 pixels of distance in the finest layer.
With reference to \cite{LefCoh01}, we can assume that on the coarsest level, the discretized energy functional 
has a unique global minimum and that the displacements are still small. We further expect to obtain the global minimum by refining 
the problem at finer scales and using the outcome of the coarser iteration as an initial guess of the fine level.

Second, this strategy results in a faster algorithm \cite{BakSchRotBlaSze11}. The optical flow is computed on the coarsest level, where the images are composed by fewest pixels, and then upsampled and used to initialize the next level. The initialization results in far lesser iterations at each level. For this reason, an algorithm using coarse-to-fine strategy tends to be significantly faster than an algorithm using only the finest level.
For the coarse-to-fine strategy, we use four pyramid levels and a bicubic interpolation 
between each level.

In this paper, we combine the coarse-to-fine strategy with two nested fixed-point iterations. 
We apply a presmoothing at each level, by convolving the images with a Gaussian kernel with standard deviation one, 
as proposed by \cite{BarFleBea94}.
We solve the minimization problem on each pyramid level starting from the coarsest one.  
There, we initialize the optical flow $\vec{u}$ by $0$. 
The solution of the minimization problem is smoothed applying a median filter, as proposed in \cite{SubRotBla14}, 
and then prolonged to the next finer level. There, we employ it for the initialization of the fixed-point iterations.  

For the purpose of numerical realization we call the iterates of the fixed point iteration 
$u_i^{(k)}$ for $k = 1,2,...,K$ where $K$ denotes the maximal number of iterations.
We plot the pseudo-code to illustrate the structure of the algorithm in Figure \ref{fig:algo}.
\begin{figure}[ht]
\begin{algorithmic}
\State Given an input sequence f
\State Initialization
\ForAll{frame in the sequence}
\State create a pyramid of 4 levels
\State calculate the saliency map
\State smooth the frame
\EndFor
\For{each level $lev \in$ {1..4} }
\If {lev=1}
	\State $u_1^{(k)}$=0
	\State $u_2^{(k)}$=0
\Else
 \State $u_1^{(k)}$= $u_1^{(k+1)}$
 \State $u_2^{(k)}$= $u_2^{(k+1)}$
	\EndIf
	\While{the precision tolerance $\geq 0.001$}
\State Calculate an approximation of $u_1^{(k+1)}$ and $u_2^{(k+1)}$
\State solving fixed point iterations
\EndWhile
\State Apply median filtering to the flow
\State Rescale $u_1^{(k+1)}$ and $u_2^{(k+1)}$ with bicubic interpolation
\EndFor
\end{algorithmic}
\caption{Pseudo-code to illustrate the structure of the 
algorithm}
\label{fig:algo}
\end{figure}

For each level of the pyramid we compute with the fixed point iterations the solution of the optimality condition of the 
functional \eqref{eq:om}: 
\begin{equation}
\begin{aligned}
\label{eq:sd}
&0=\sum_{i=1}^\sigma \mathcal{B}_{i} \frac{\partial f_i}{\partial x_1} \left(\frac{\partial f_i}{\partial x_1} u_1 +
\frac{\partial f_i}{\partial x_2} u_2 +\frac{\partial f_i}{\partial t} \right)
-\nabla_3 \cdot (\Psi^\prime (|\nabla_3 u_1|^2+|\nabla_3 u_2|^2)\nabla_3 u_1)\\
&0=\sum_{i=1}^\sigma \mathcal{B}_{i} \frac{\partial f_i}{\partial x_2} \left(\frac{\partial f_i}{\partial x_1} u_1 +
\frac{\partial f_i}{\partial x_2} u_2+\frac{\partial f_i}{\partial t}\right)-\nabla_3 \cdot (\Psi^\prime (|\nabla_3 u_1|^2+|\nabla_3 u_2|^2)\nabla_3 u_2)
\end{aligned}
\end{equation} 
where $\mathcal{B}_{i}$ is the \emph{i-component} of the vector $\mathcal{B}$ and $\sigma=2,4$,  
if we consider intensity with complemented saliency data or color with complemented saliency data, respectively. 

For the solution of the system of equations we use a semi-implicit Euler method: Let $\tau$ be the step size, then 
the fixed point iterations is defined by
\begin{equation}
\begin{aligned}
\label{eq:sdi}
\frac{u_1^{(k+1)}-u_1^{(k)}}{\tau} & =-\sum_{i=1}^\sigma \mathcal{B}_i \frac{\partial f_i}{\partial x_1}
\left( \frac{\partial f_i}{\partial x_1} u_1^{(k+1)}+\frac{\partial f_i}{\partial x_2} u_2^{(k)}+
       \frac{\partial f_i}{\partial t} \right)+\\
&\quad \nabla_3 \cdot (\Psi^\prime (|\nabla_3 u_1^{(k)}|^2+|\nabla_3 u_2^{(k)}|^2)\nabla_3 u_1^{(k)})\\
\frac{u_2^{(k+1)}-u_2^{(k)}}{\tau}&=-\sum_{i=1}^\sigma \mathcal{B}_i \frac{\partial f_i}{\partial x_2} 
\left(\frac{\partial f_i}{\partial x_1} u_1^{(k)}+\frac{\partial f_i}{\partial x_2} u_2^{(k+1)}+
\frac{\partial f_i}{\partial t}\right)\\
&\quad +\nabla_3 \cdot (\Psi^\prime (|\nabla_3 u_1^{(k)}|^2+|\nabla_3 u_2^{(k)}|^2)\nabla_3 u_2^{(k)})\\
\end{aligned}
\end{equation}
where for the discretization of $\nabla_3 \cdot (\Psi^\prime (|\nabla_3 u_1^{(k)}|^2+|\nabla_3 u_2^{(k)}|^2)\nabla_3 u_1^{(k)})$ 
and $\nabla_3 \cdot (\Psi^\prime (|\nabla_3 u_1^{(k)}|^2+|\nabla_3 u_2^{(k)}|^2)\nabla_3 u_2^{(k)})$ we follow \cite{WeiSchn01b}.

In our experiments we use $\tau=10^{-3}$. Moreover, we set the regularization parameter $\alpha=40$ and $\lambda=1$, 
unless stated otherwise.
The iterations are stopped, when the Euclidean norm of the relative error
$$\frac{|u_j^{(k)}-u_j^{(k+1)}|}{|u_j^{(k)}|}, \hspace{1cm} j=1,2$$ 
drops below the precision tolerance value of $tol=0.003$ for both the components $u_j$.
For the discretization of \eqref{eq:sdi}, we use central difference approximations of $\frac{\partial f_i}{\partial x_1},\frac{\partial f_i}{\partial x_2},\frac{\partial f_i}{\partial t}$ for each pixel.
We consider that the spacing in the central differences approximations equals a value of one with reference to space and time.

\section{Eye tracking experiment}
\label{sec:eyeTrack}
We next test our model. This is
done by recording human participants' fixations on small video clips of relatively total views of natural
scenes, and testing how much of variance of fixation locations could be explained by our model as
compared to two alternative models, \cite{GuoZha10} and \cite{SubRotBla14}. To this end, model
performances are compared using the areas under the curves (AUCs) and normalized scanpath
saliency
\subsubsection{Participants}{Twenty-four (five female) human viewers with a mean age of 25 years (range 19--32) 
volunteered in an eye tracking experiment and received partial course credit in exchange. All were undergraduate 
Psychology students at the University of Vienna.  Viewers were pre-screened for normal or fully corrected eye-sight 
and intact color vision. Prior to the start of the experiment, written informed consent was obtained from all participants. }

\subsubsection{Stimuli}{The same 71 short video recordings that were used for performing the saliency computations 
were presented to the sample of human viewers. All videos were presented without sound. Each of the videos contained 
moving and potentially interesting content at several spatially distinct locations off-center (e.g., moving cars, wind 
moving trees, people crossing streets etc.). 
Hence, videos presented viewers with multiple potentially
interesting fixation locations and the viewers could visually explore the scene.
This distinguished the videos used here from professionally produced 
footage from TV shows or feature films, which often elicit strong 
tendencies to keep fixation at the center of the movie scene \cite{dorr2010variability}. The order in which the 
videos were presented in the experiment was chosen randomly for each participant. All videos were presented in full 
screen on a CRT monitor. 
}

\subsubsection{Apparatus}{Throughout each data acquisition session, the viewer’s dominant eye 
position was recorded using an EyeLink 1000 Desktop Mount (SR Research Ltd., Kanata, ON, Canada) video-based eye 
tracker sampling at 1000 Hz. The eye tracker was calibrated using a standard 9-point calibration sequence. 
Prior to each individual video, a fixation circle was presented at the center of the screen to perform a drift check. 
Whenever the acquired gaze position differed by more than $1^{\circ}$ from the fixation target's position, the whole 
calibration sequence was repeated to assure maximal spatial accuracy for each viewer. Video stimuli were delivered in 
color to a 19-in. CRT monitor (Multiscan G400, Sony Inc.) at a screen resolution of $1280 \times 1024$ pixels 
(85 Hz refresh rate). Viewers sat in front of the monitor and placed their head on a chin and forehead rest, which 
held viewing distance fixed at 64 cm, resulting in an apparent size of each full-screen video of $31 \times 24.2^{\circ}$. 
The presentation procedure was implemented in MATLAB with the PsychToolbox and the Eyelink toolbox functions 
\cite{brainard1997psychophysics,cornelissen2002eyelink,pelli1997videotoolbox}. 
}

\subsubsection{Data preprocessing}{For the evaluation of the model results we compared the spatial distribution 
of the human viewers' fixations on each video frame with the computed dynamic saliency maps for each frame. The 
recorded gaze position vector was parsed into three classes of oculomotor events: blinks, saccades, and fixations. 
Fixations were defined as the mean horizontal and vertical gaze position during data segments not belonging to a blink, 
or a saccade (gaze displacement $< 0.1^\circ$, velocity $<30^\circ/s$, and acceleration $<8,000^\circ/s^2$). The parsed fixations were mapped onto each video frame depending on their start and end times. For example, if a fixation started 1.25 s after the onset the scene at location (x,y), this location was marked as fixated in a fixation matrix belonging to the 30th frame of the video. If this same fixation ended 2 s after the onset of the video, the corresponding fixation matrix from the 30th through to the 50th frame were set to true (or fixated) at that location. This mapping was done for all viewers, and all videos, resulting in a 3-dimensional fixation matrix with the spatial resolution of one video frame and the temporal extent of the number of analyzed video frames. Each video was presented for 10 s during the data collection.
}
\section{Results}
\label{sec:results}

\subsection{Qualitative model evaluation}

\begin{figure}[t]
\begin{center}
\includegraphics[scale=0.196]{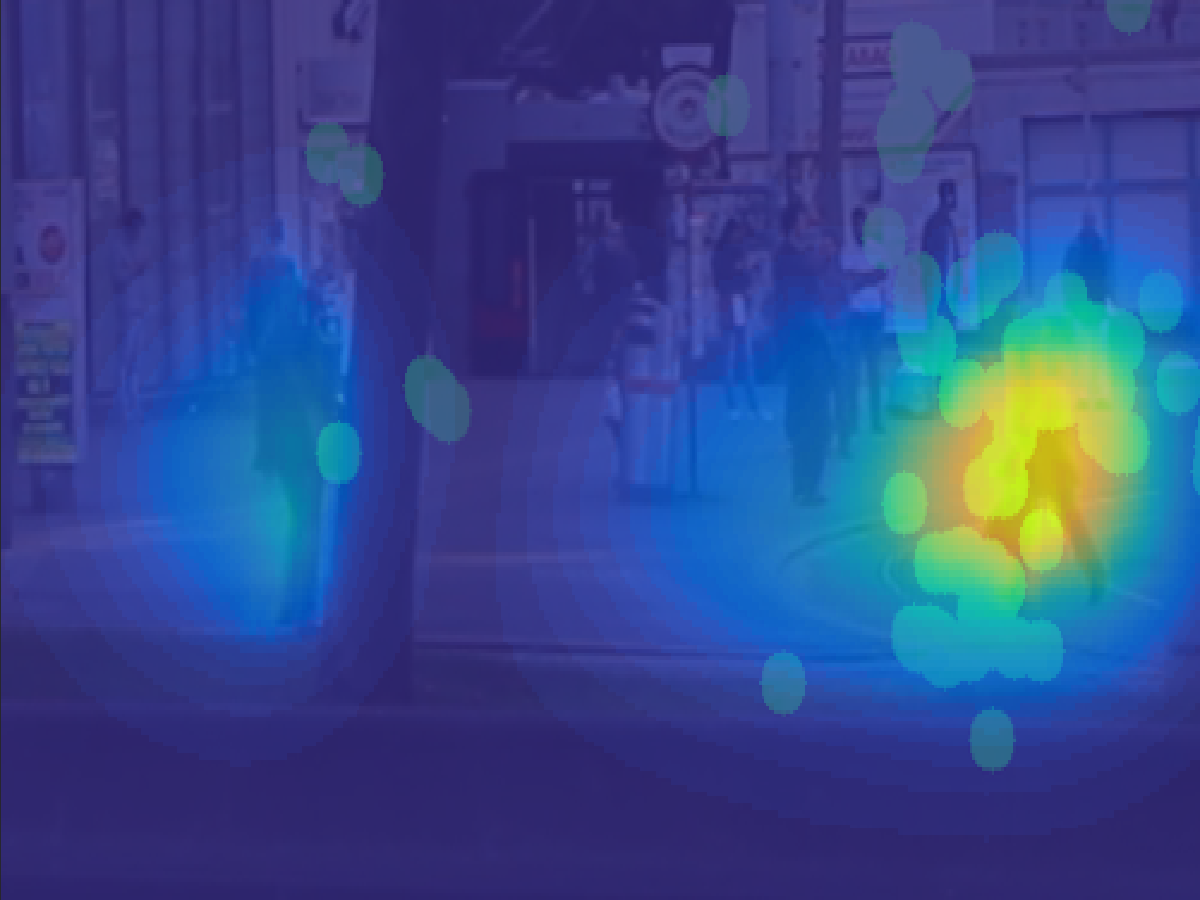} 
\includegraphics[scale=0.196]{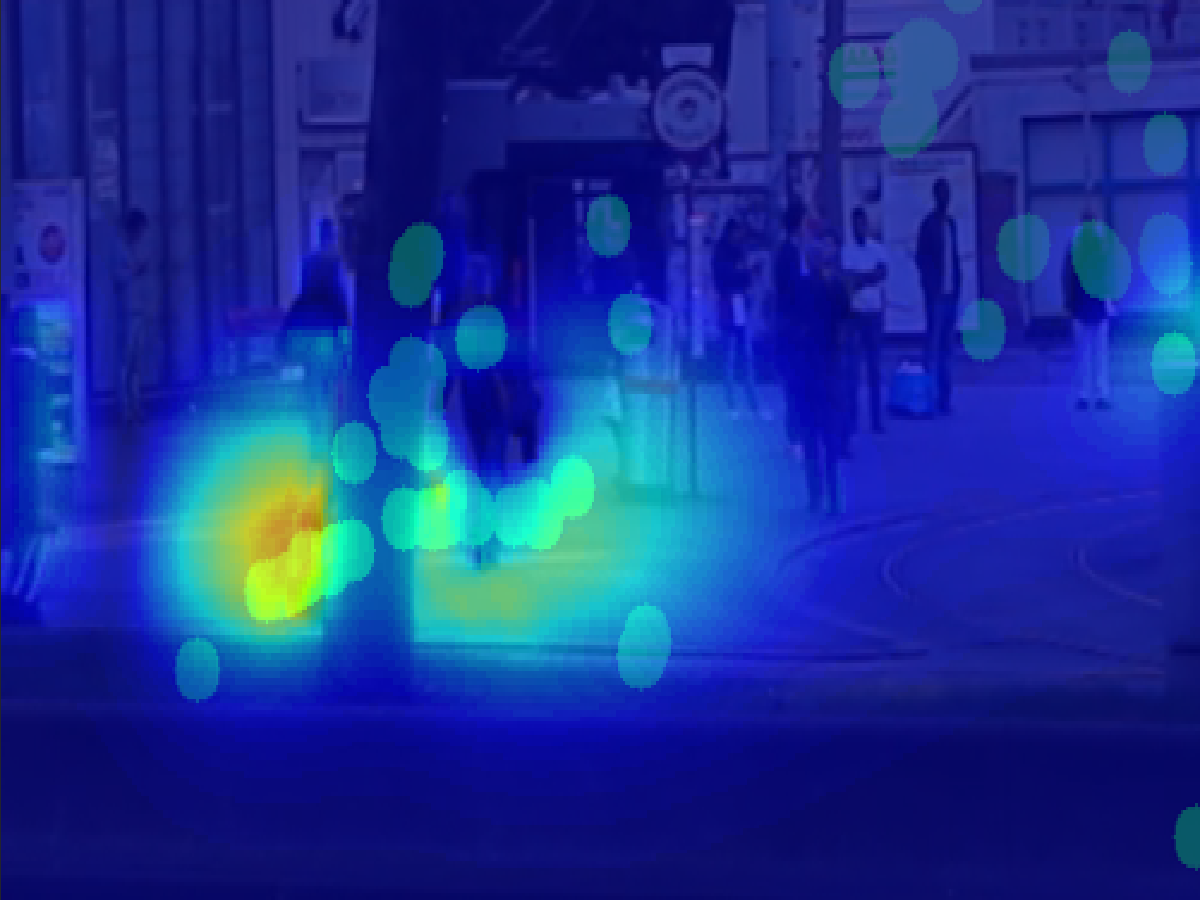} 
\includegraphics[scale=0.196]{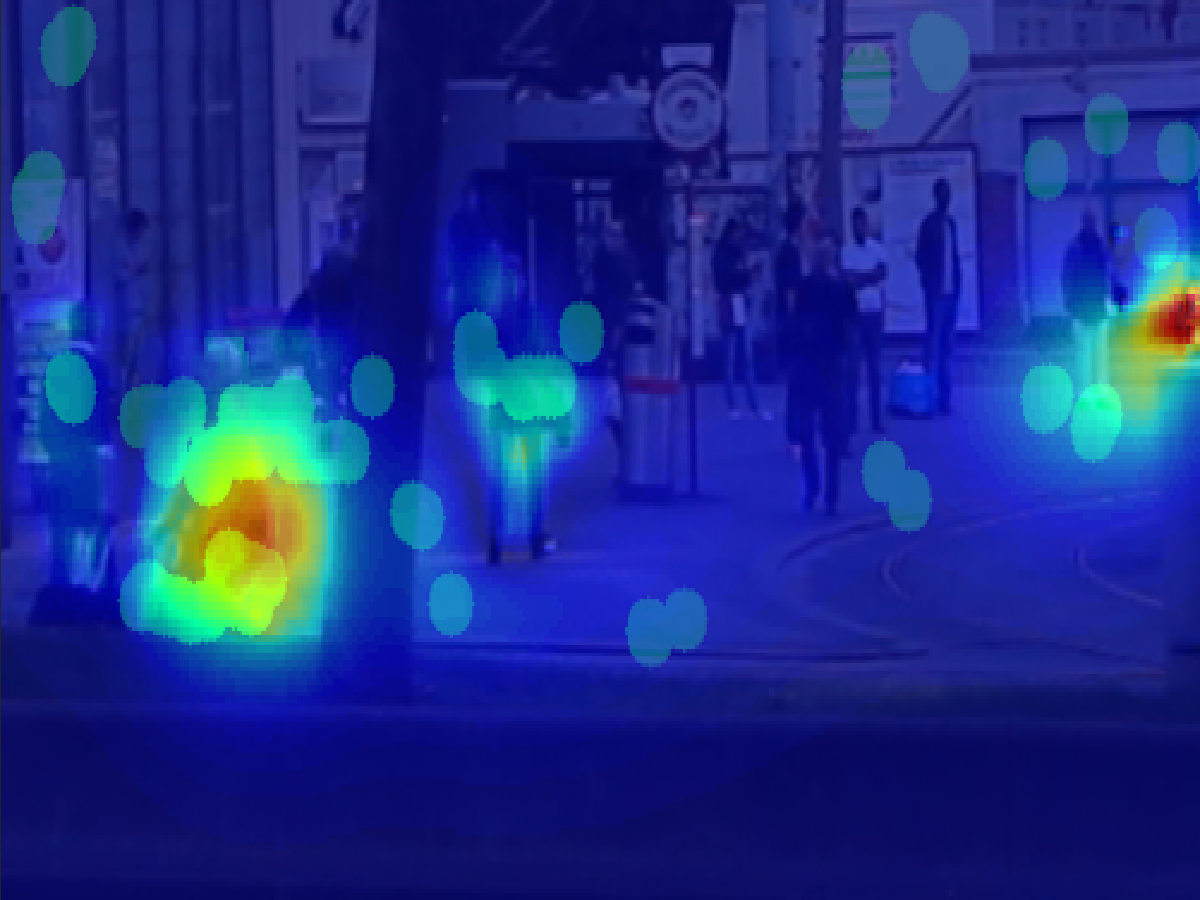}\\ 
\includegraphics[scale=0.196]{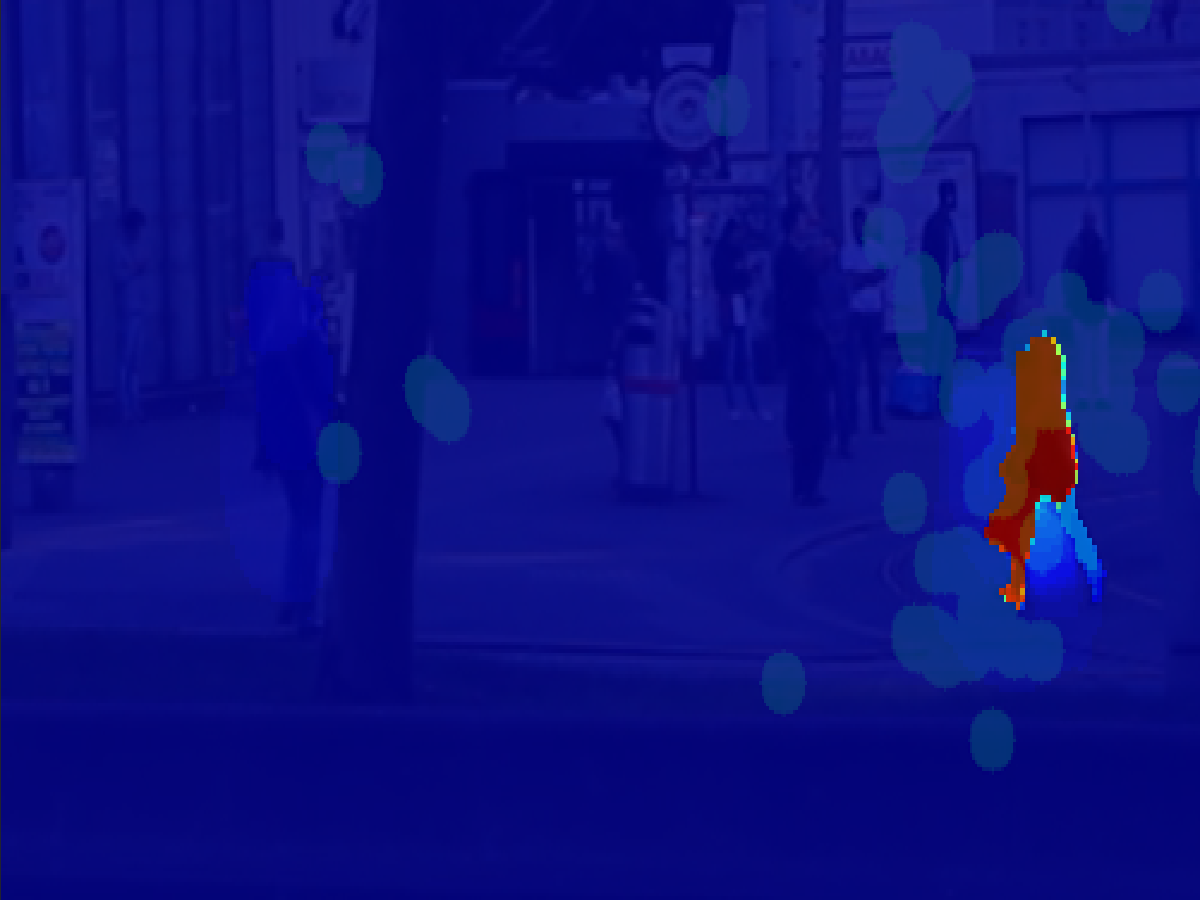} 
\includegraphics[scale=0.196]{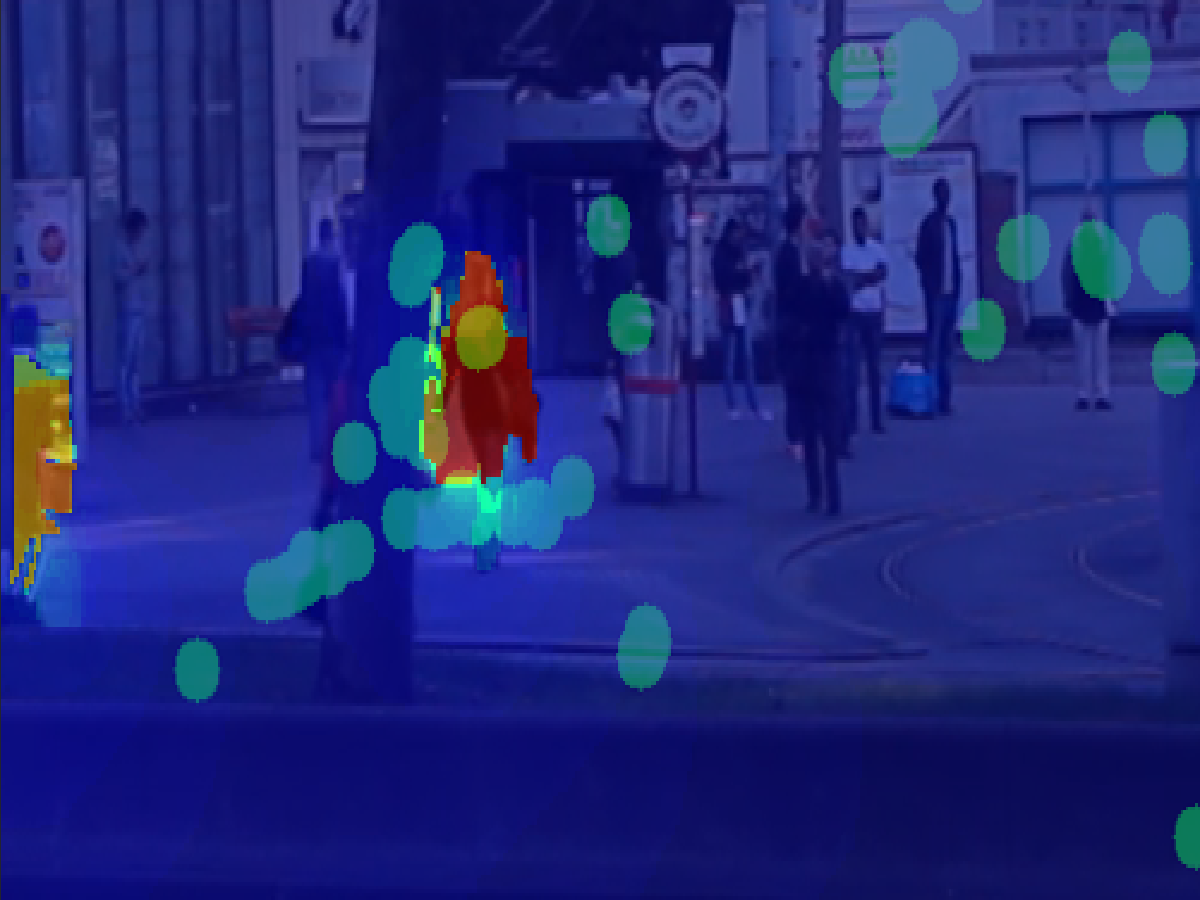} 
\includegraphics[scale=0.196]{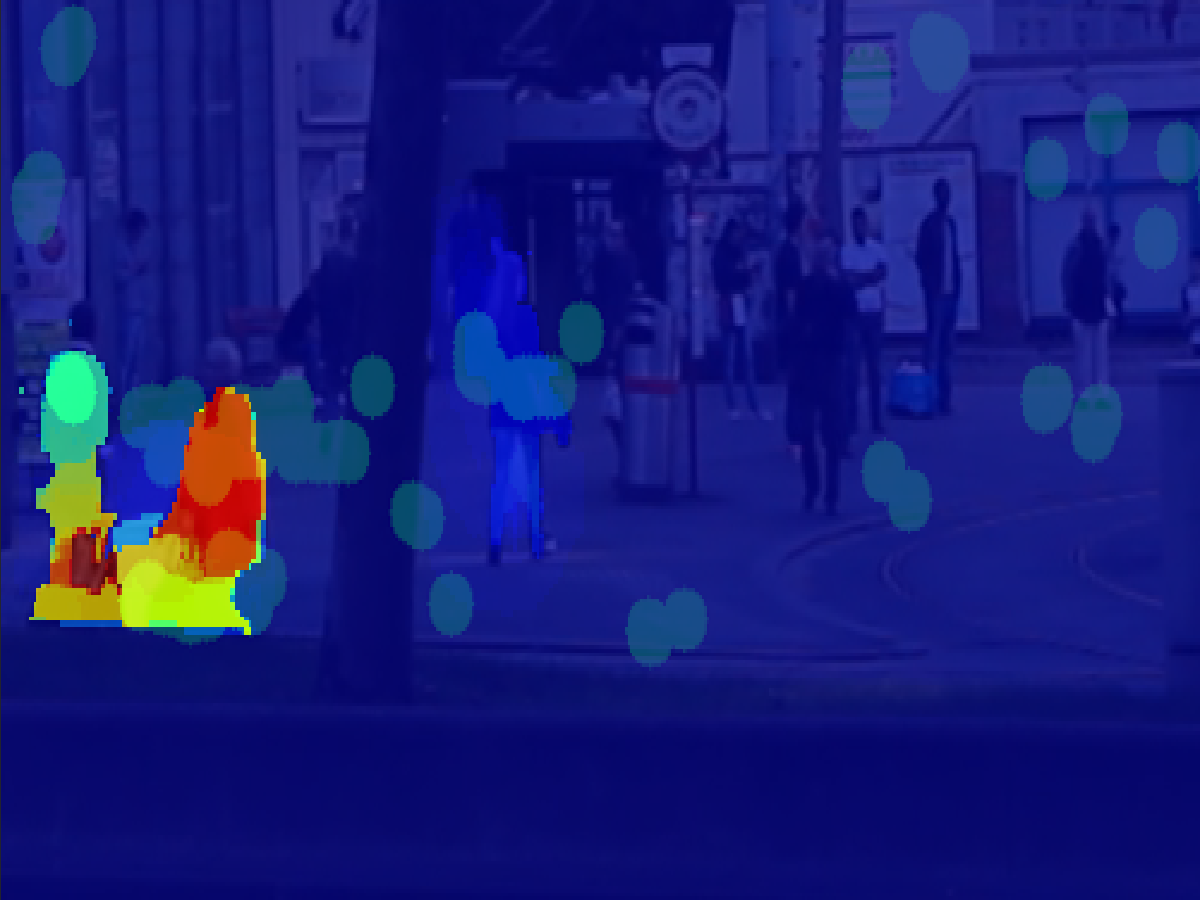}\\
\includegraphics[scale=0.196]{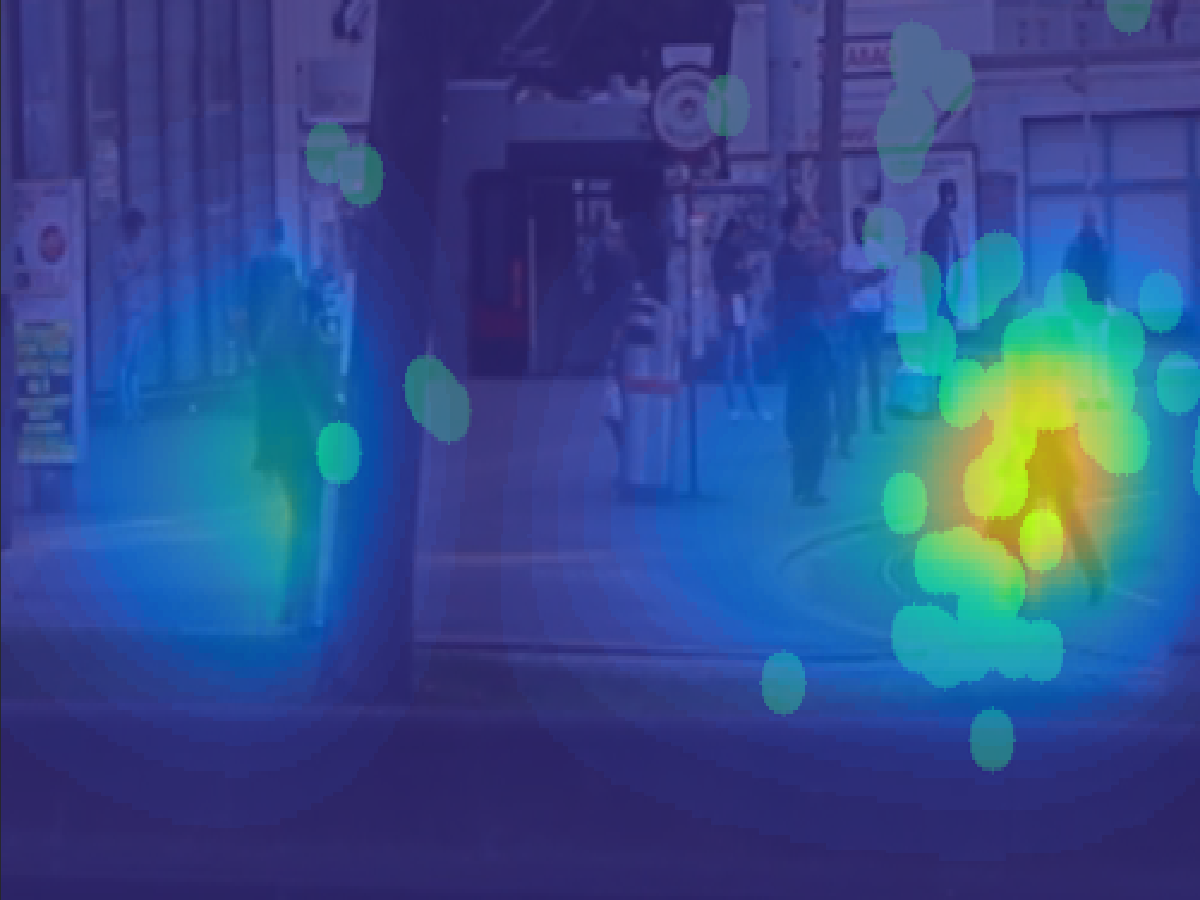} 
\includegraphics[scale=0.196]{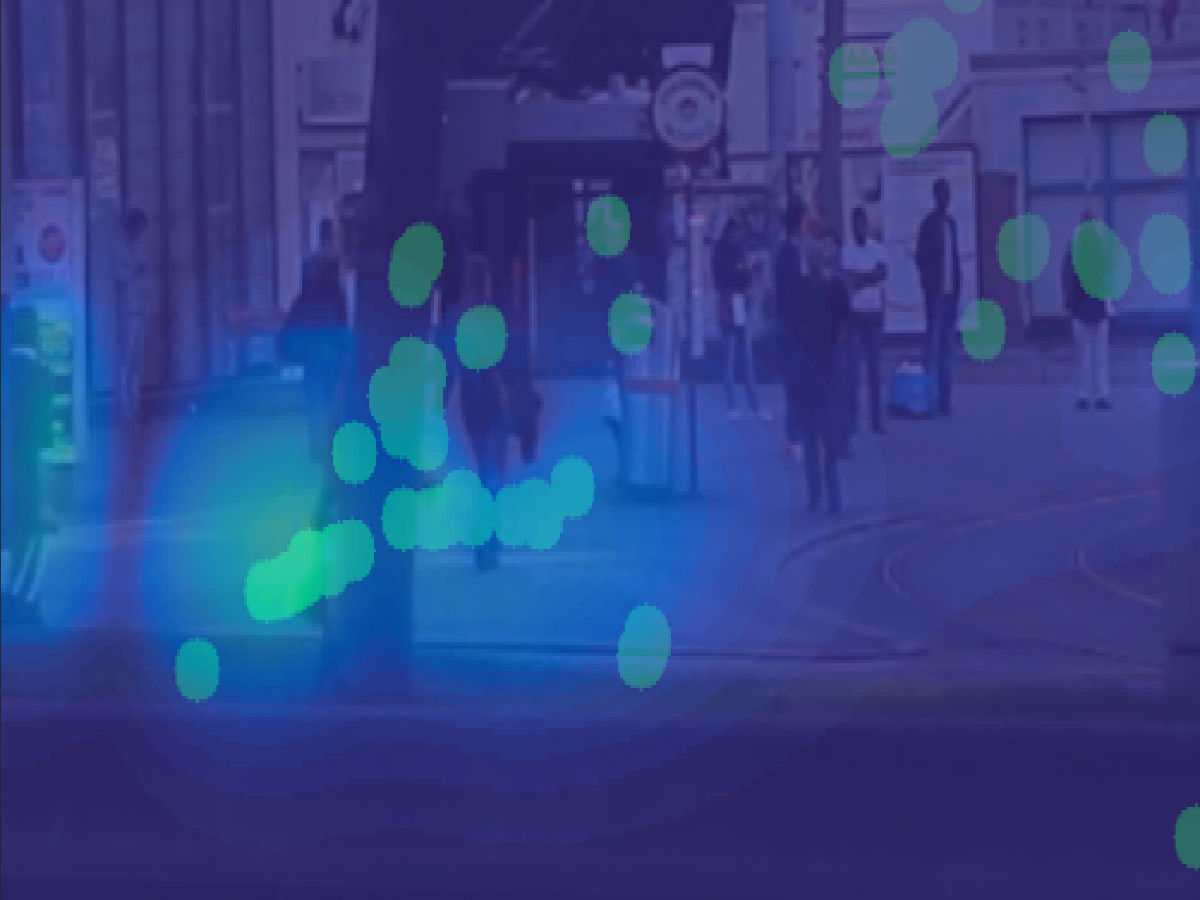} 
\includegraphics[scale=0.196]{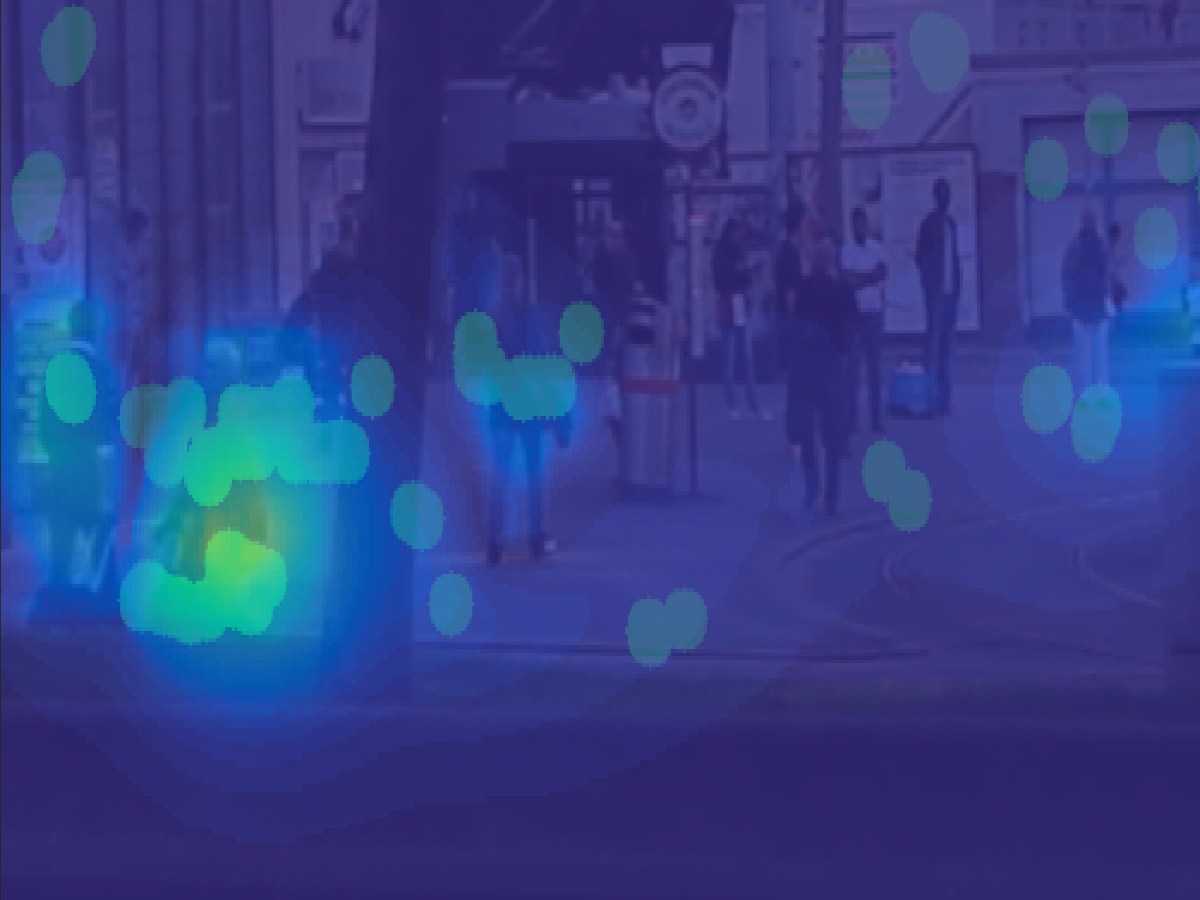} 
\end{center}
\caption{
The dynamic sequence shows people walking in front of a metro station. A woman in a red coat is walking behind a 
tree and is salient. 
We notice that when the woman is occluded (central column), only \eqref{eq:om} with saliency complemented data (top) is 
able to recognize the occluded part as salient. The method considering only two frames \protect\cite{SubRotBla14} (middle) 
does not recognize this area as salient. 
The method \eqref{eq:om} without complemented data (bottom) results in a lower correlation between the saliency map and the fixation distribution.}
\label{fig:exp5050}
\end{figure}

\begin{figure}[t]
\begin{center}
\includegraphics[scale=0.196]{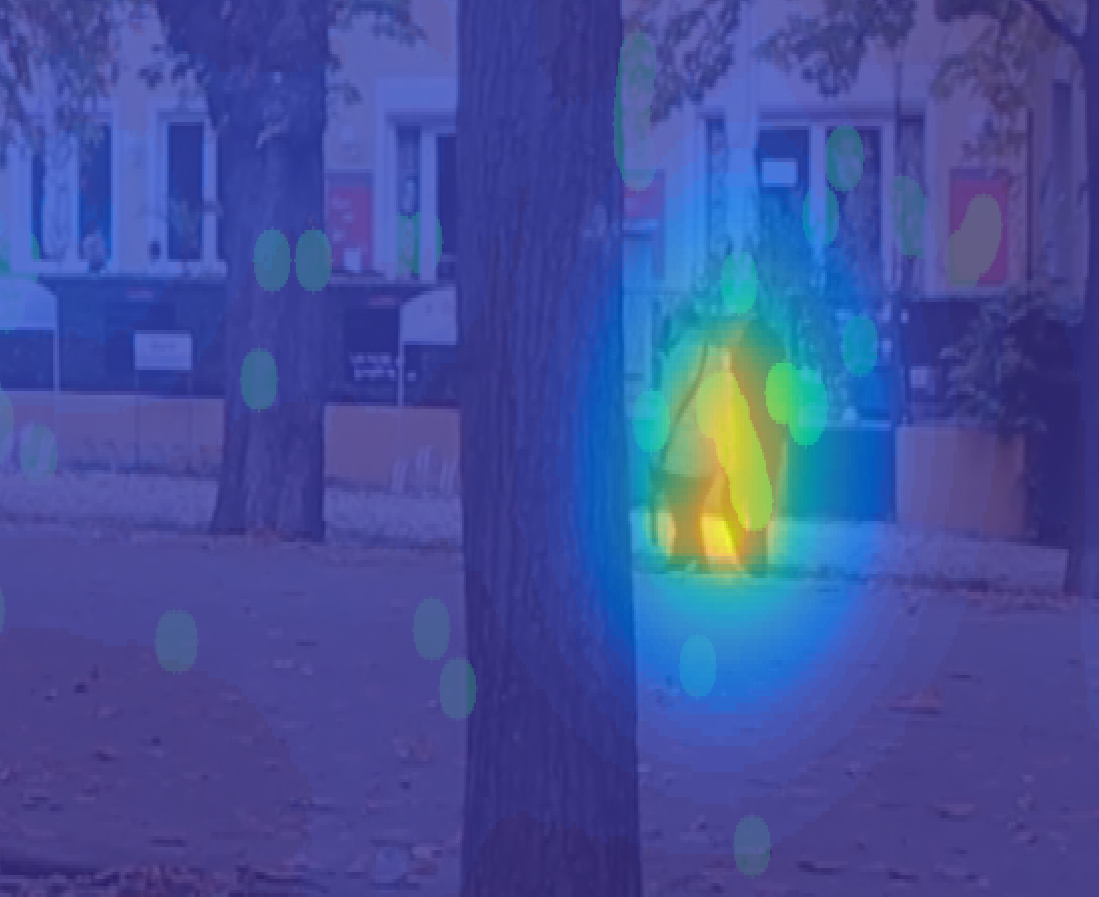} 
\includegraphics[scale=0.196]{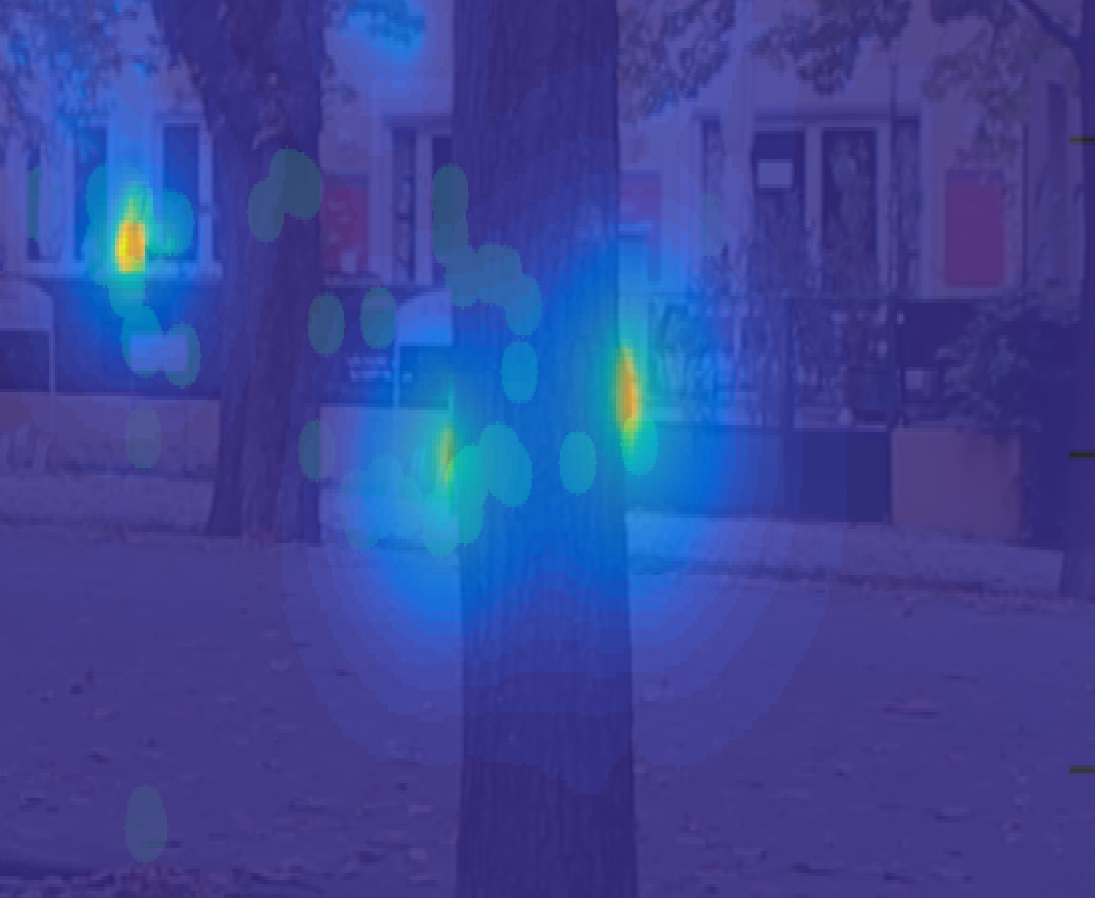} 
\includegraphics[scale=0.196]{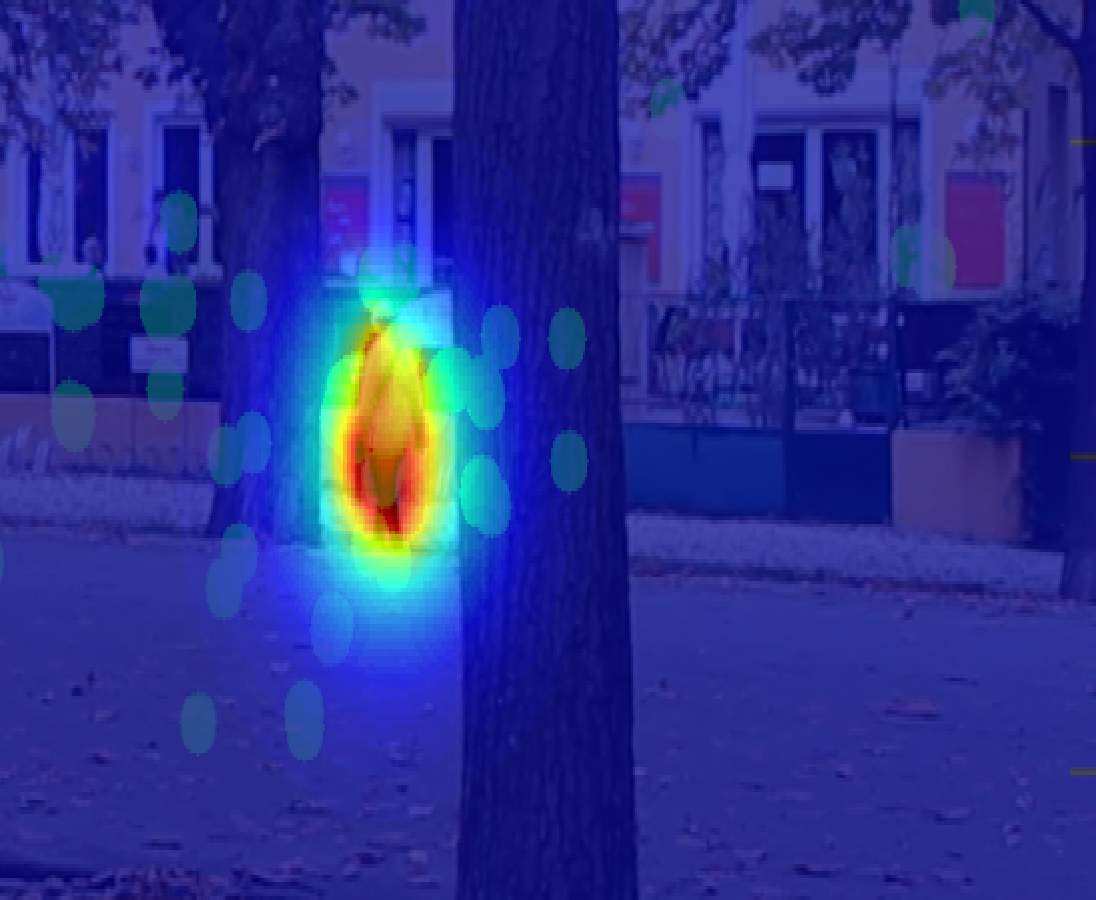}\\ 
\includegraphics[scale=0.196]{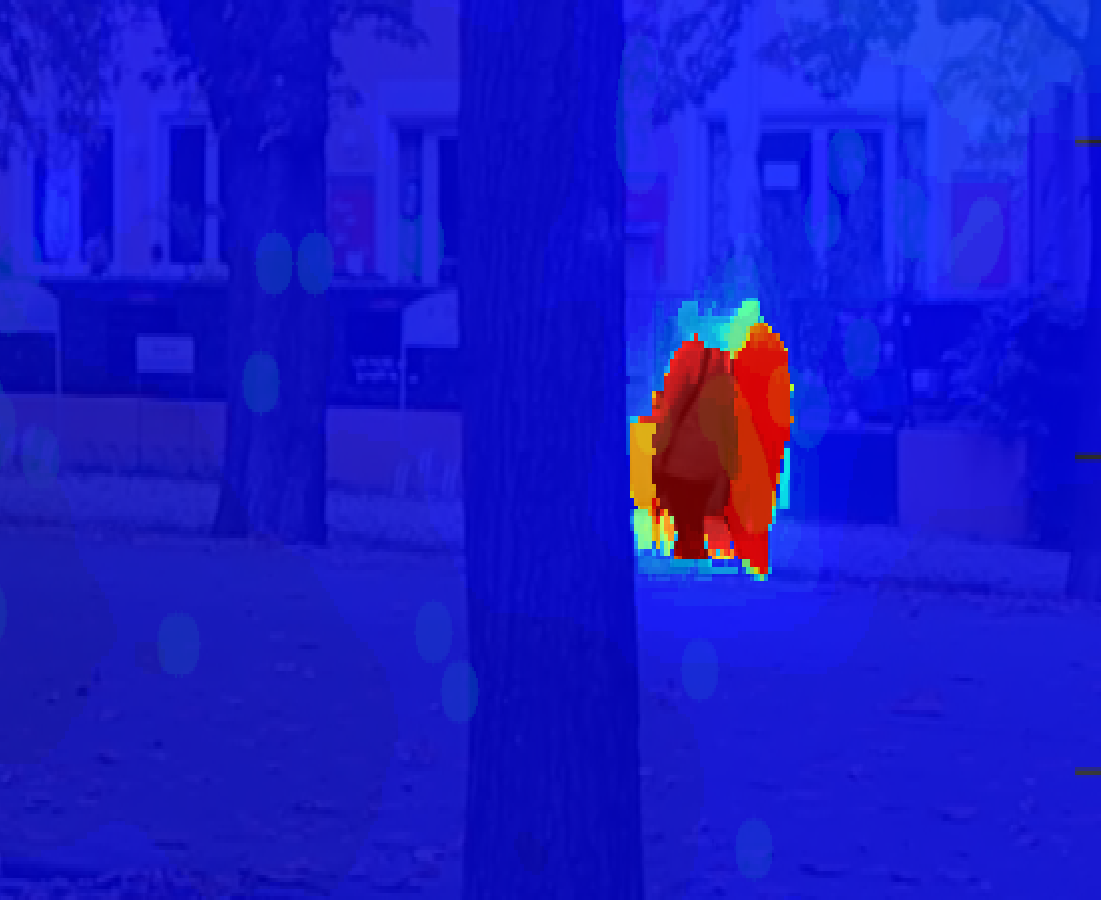} 
\includegraphics[scale=0.196]{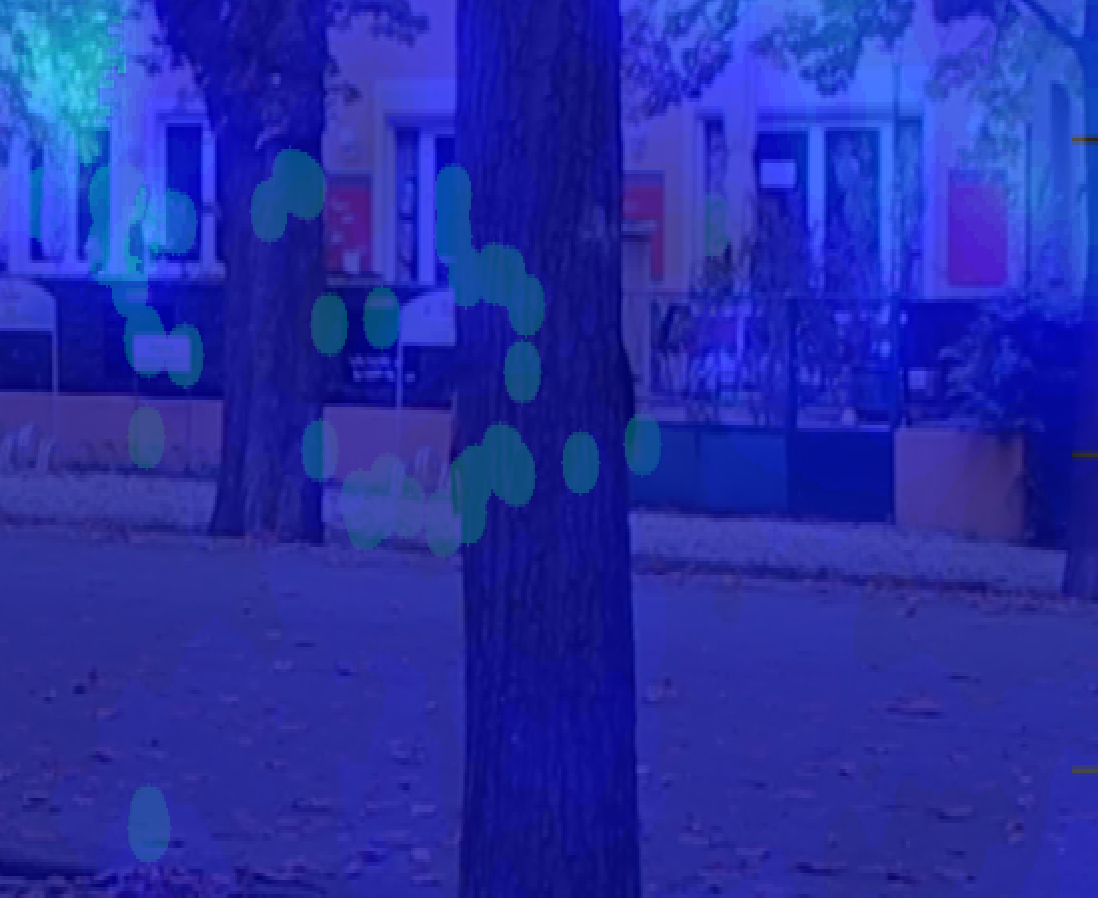} 
\includegraphics[scale=0.196]{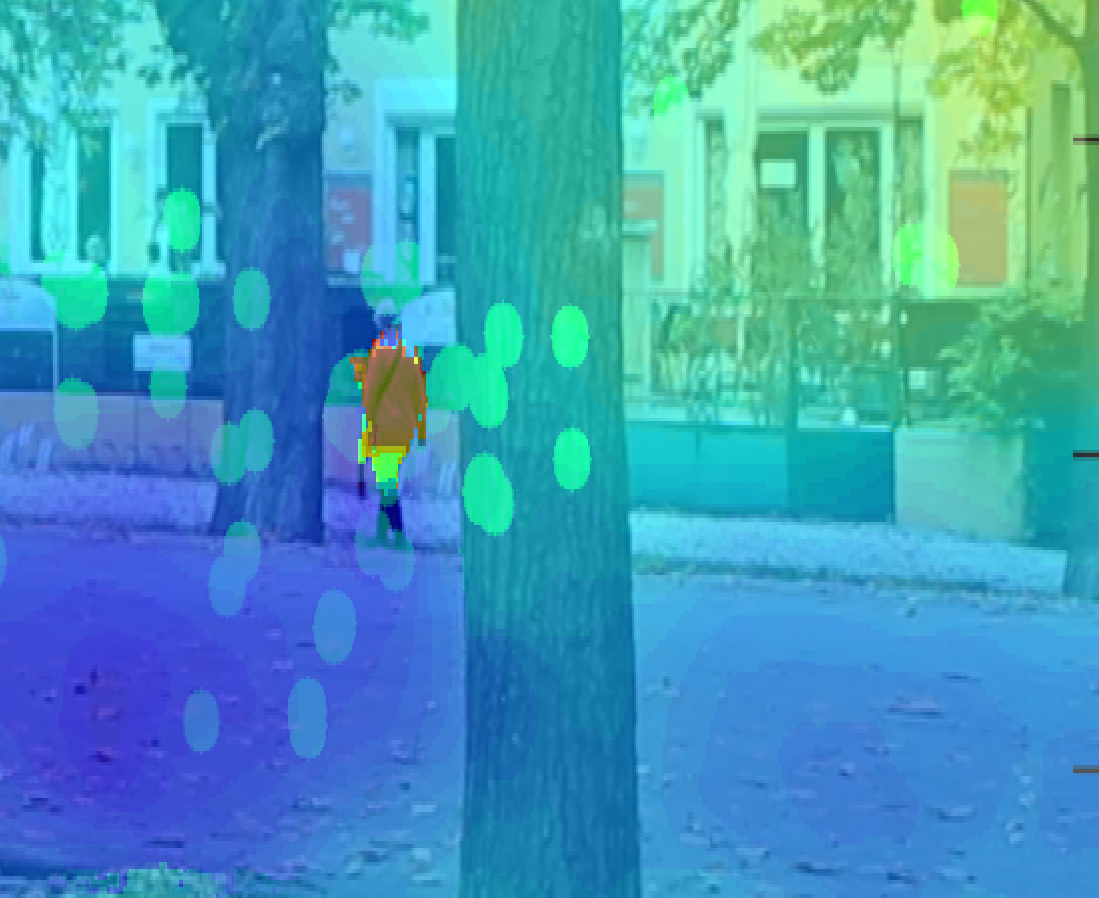}\\
\includegraphics[scale=0.196]{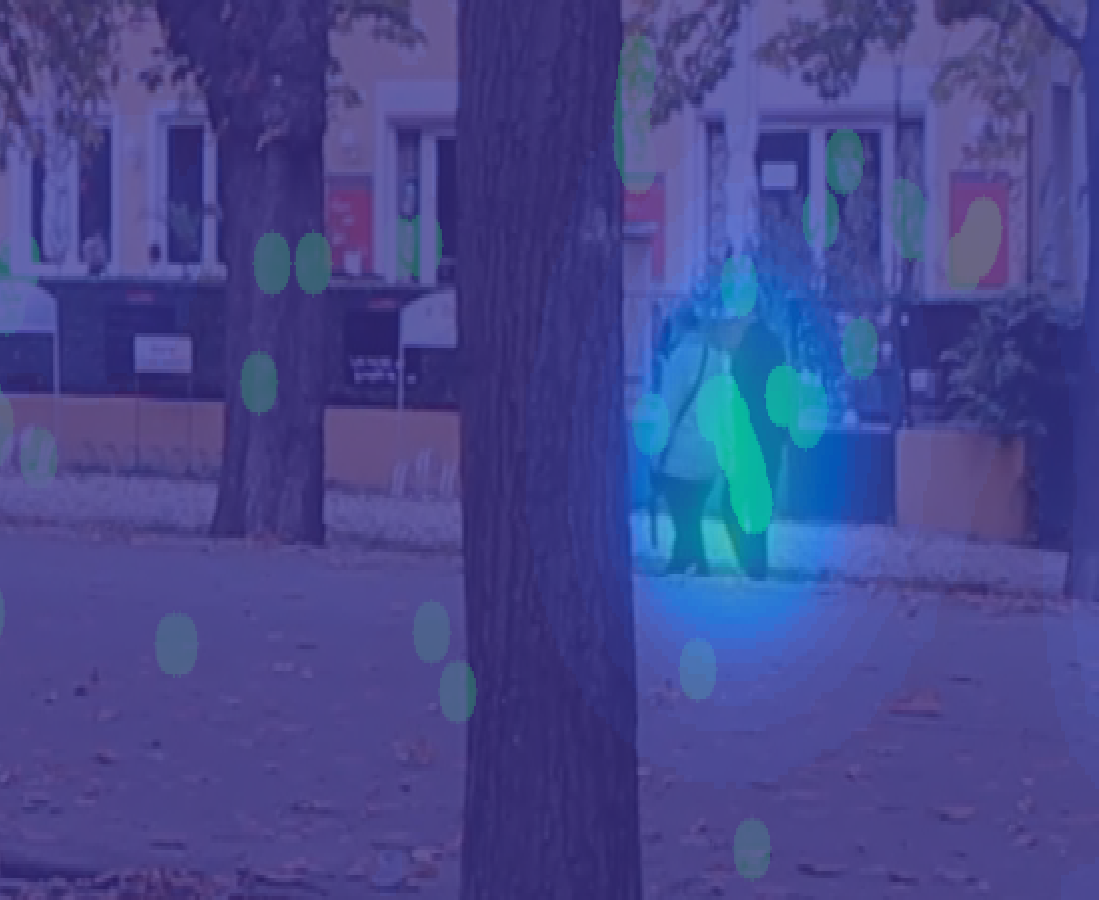} 
\includegraphics[scale=0.196]{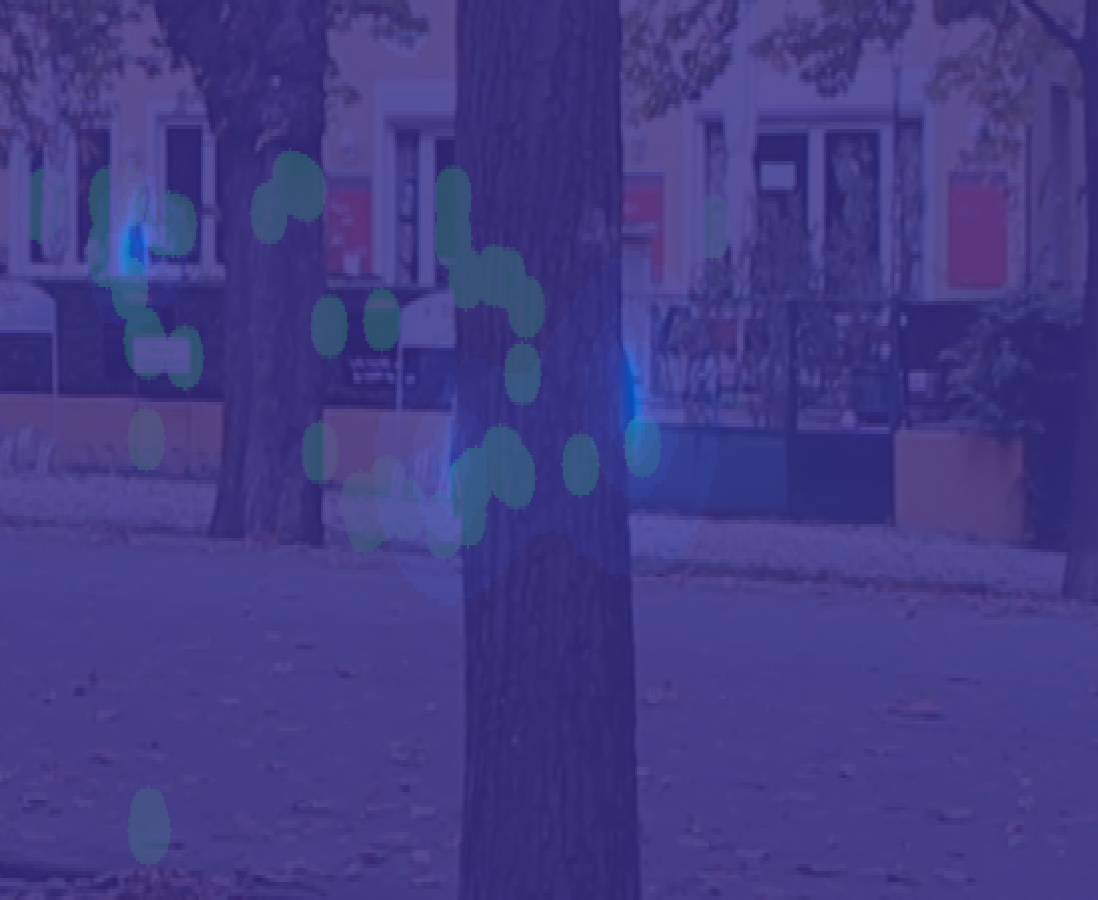} 
\includegraphics[scale=0.196]{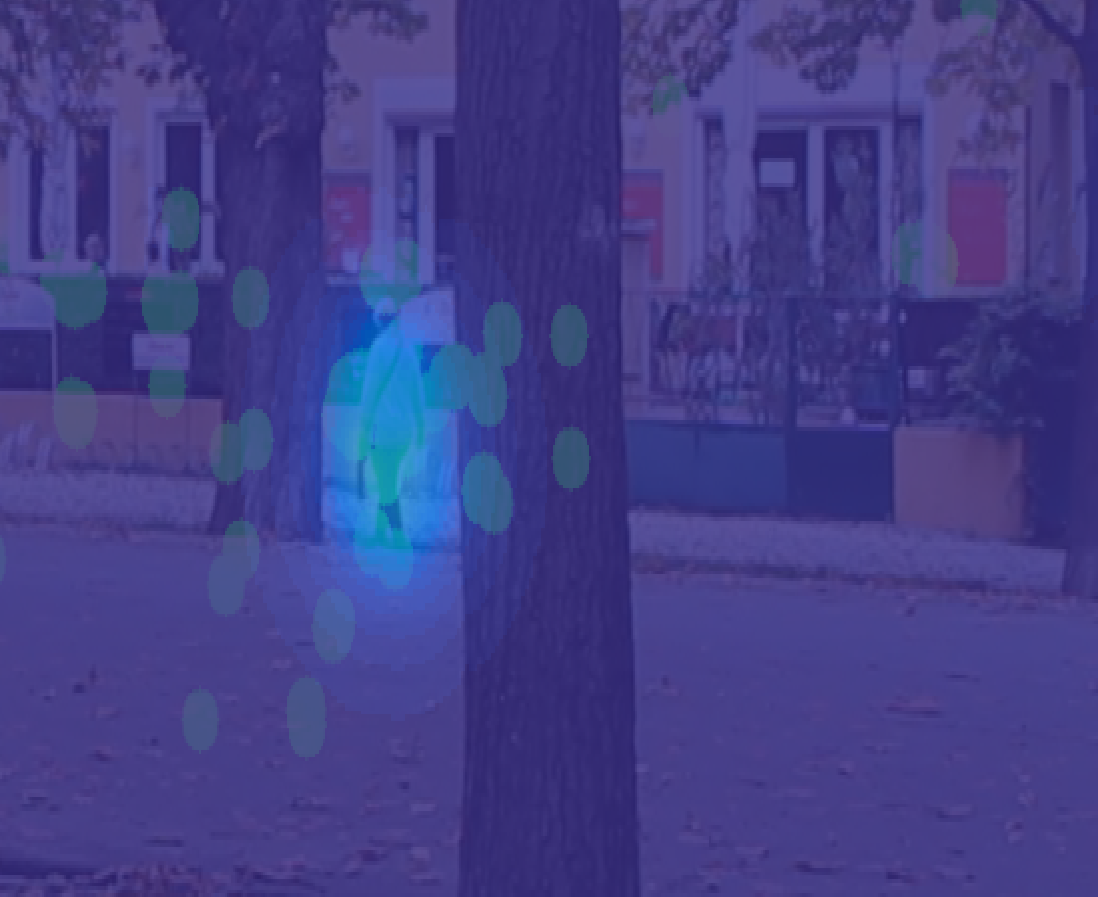} 
\end{center}
\caption{
The dynamic sequence shows a couple walking in a park. They are walking behind a tree. 
We notice that when the couple is occluded (central column), only \eqref{eq:om} with saliency complemented data (top) is 
able to recognize the occluded part as salient. The method considering only two frames \protect\cite{SubRotBla14} (middle) 
does not recognize this area as salient. The method \eqref{eq:om} without complemented data (bottom) results in a lower correlation between the saliency map and the fixation distribution. This sequence is particularly challenging due to light reflections and noise. 
We notice how the method in  \protect\cite{SubRotBla14} (middle) is affected by over-smoothing.}
\label{fig:exp5121}
\end{figure}

This section is devoted to the evaluation of the dynamical saliency mapping by comparison with eye tracking data. 
High spatial saliency should correspond to active visual attention.
Particular emphasis is put on the participants' tracking of moving objects which are
temporarily occluded, because this is a situation where standard optical flow algorithms fail although
humans try to actively track such objects \cite{AlvHorArsDimWol05, FloSchoPyl08, SchoPyl99} that is, they attend to such temporarily occluded objects.
%
%
\begin{figure}[t]
\includegraphics[scale=0.196]{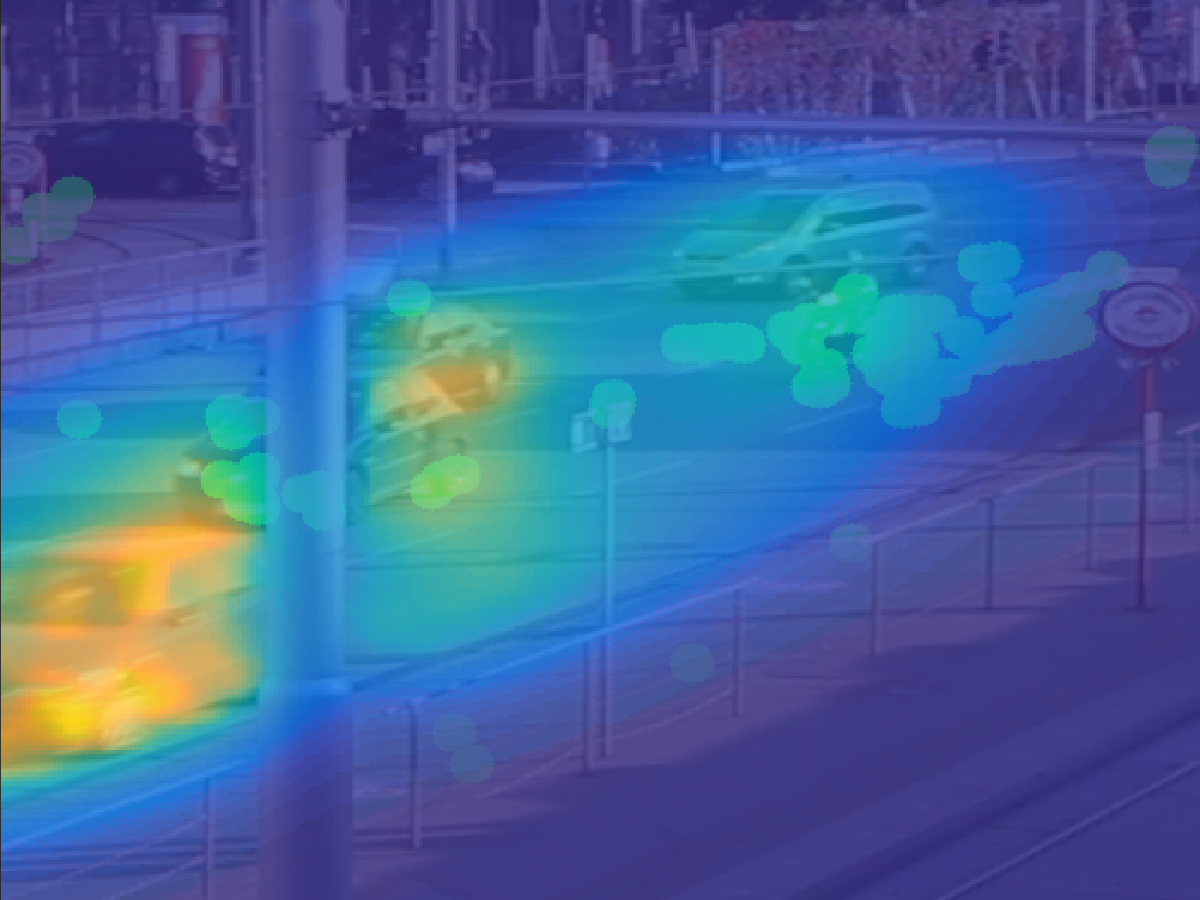} 
\includegraphics[scale=0.196]{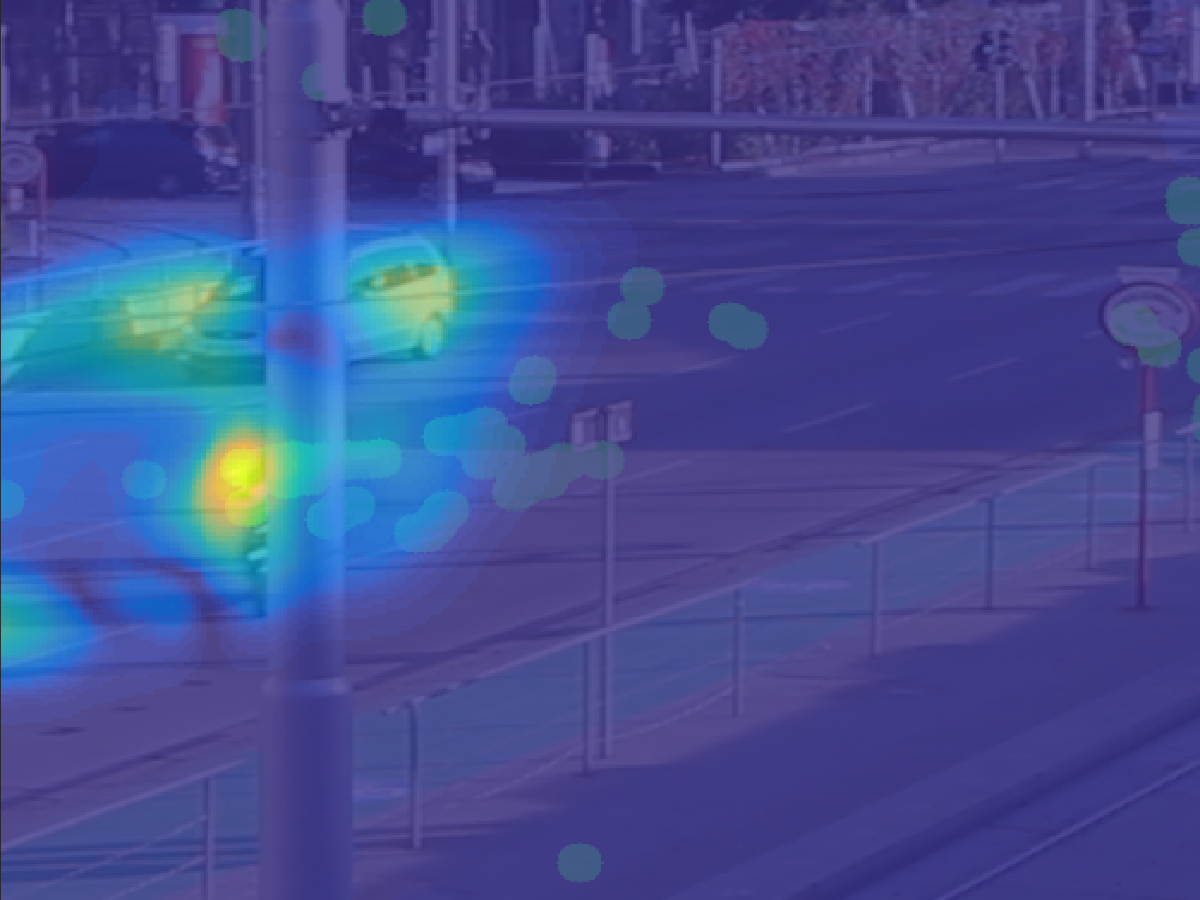} 
\includegraphics[scale=0.196]{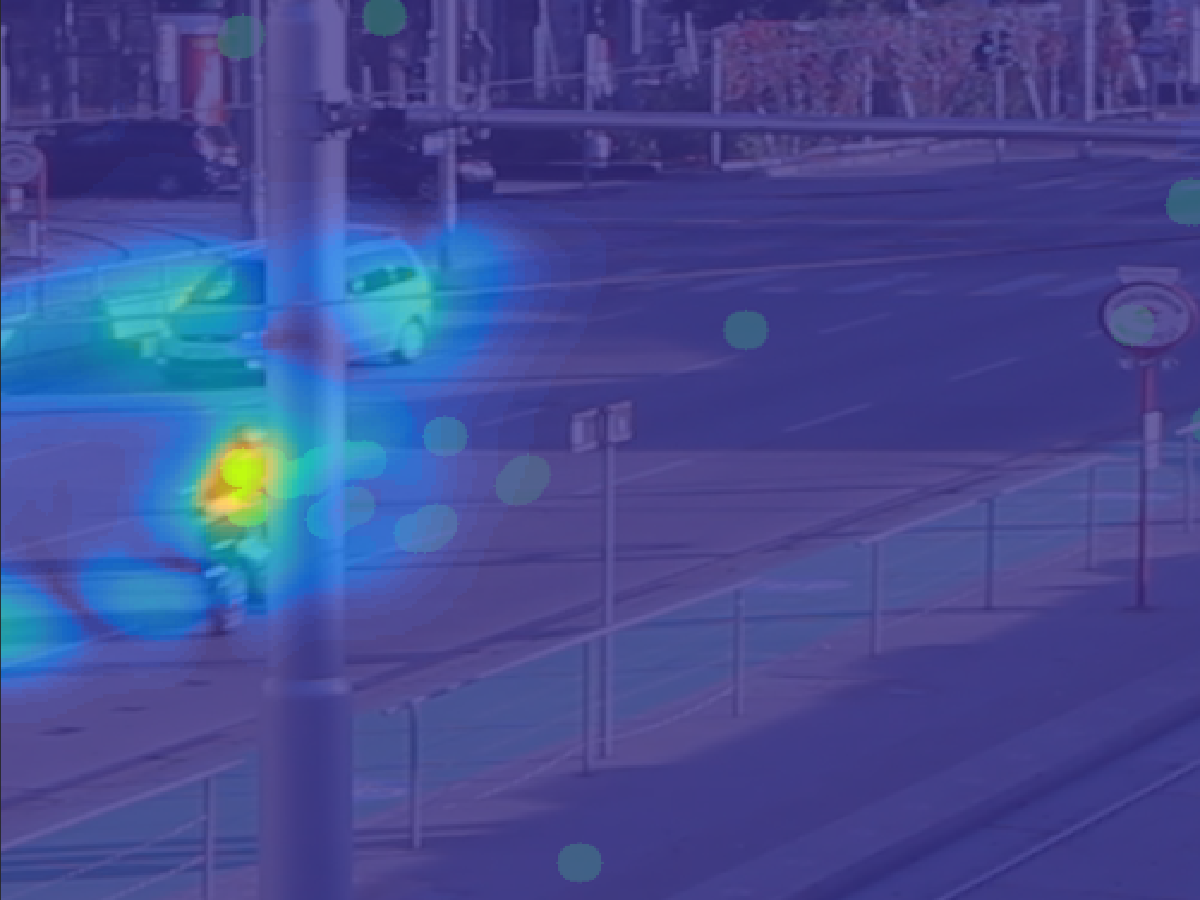}\\ 
\includegraphics[scale=0.196]{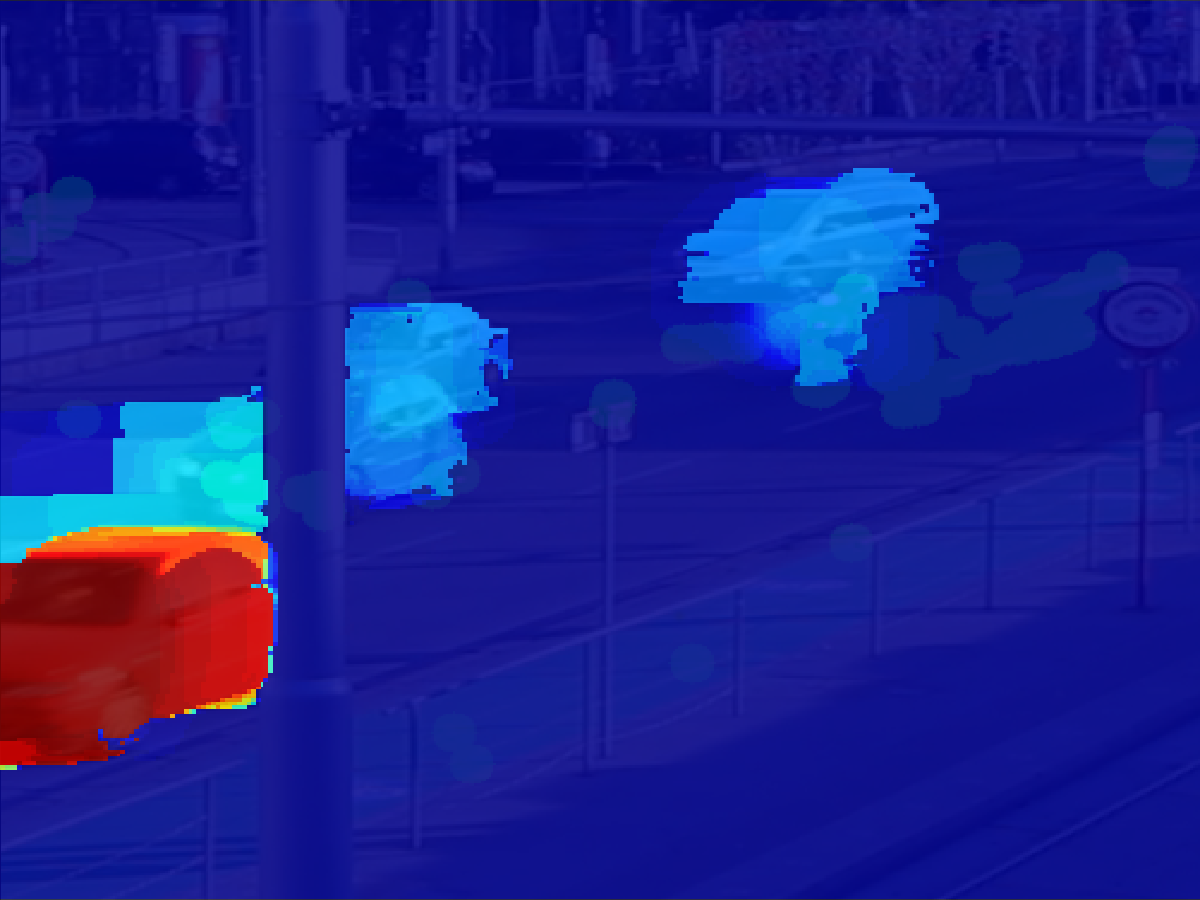} 
\includegraphics[scale=0.196]{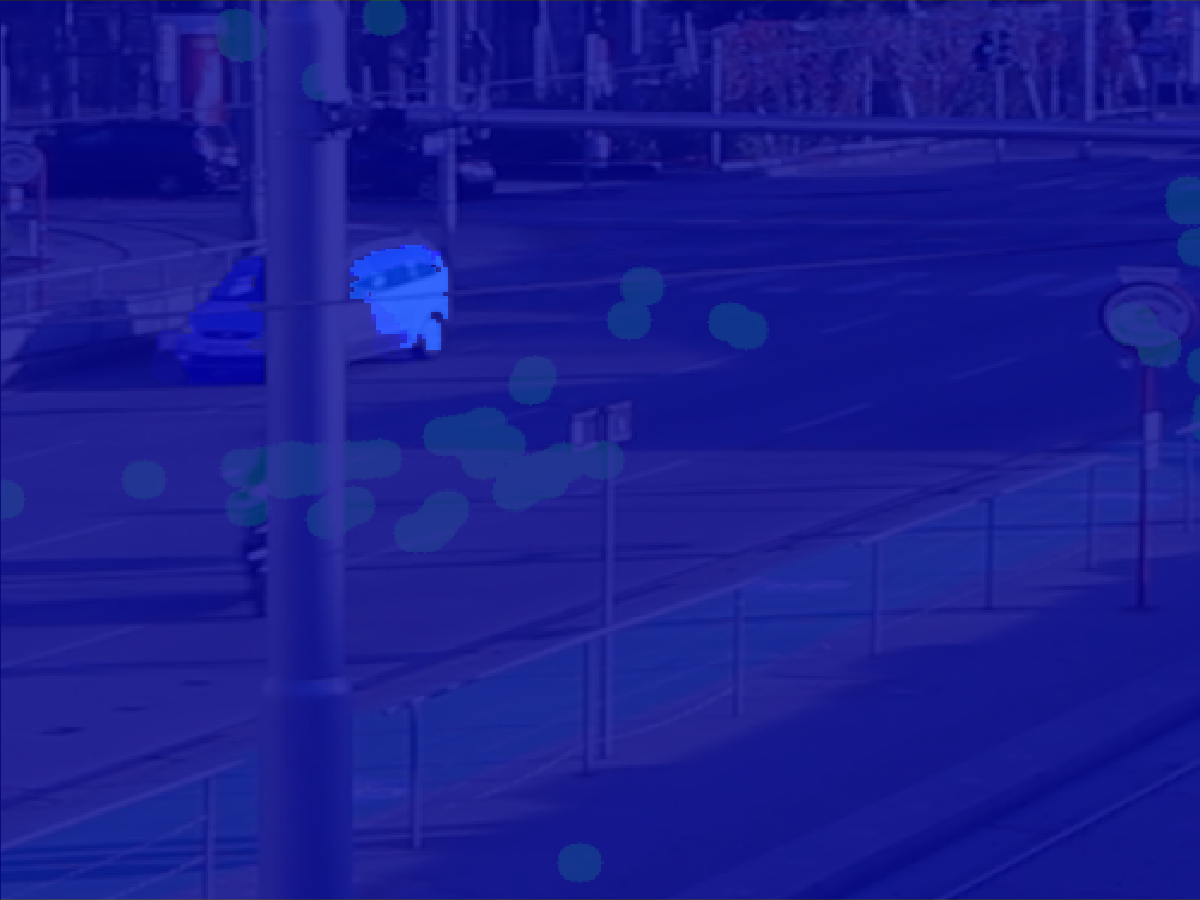} 
\includegraphics[scale=0.196]{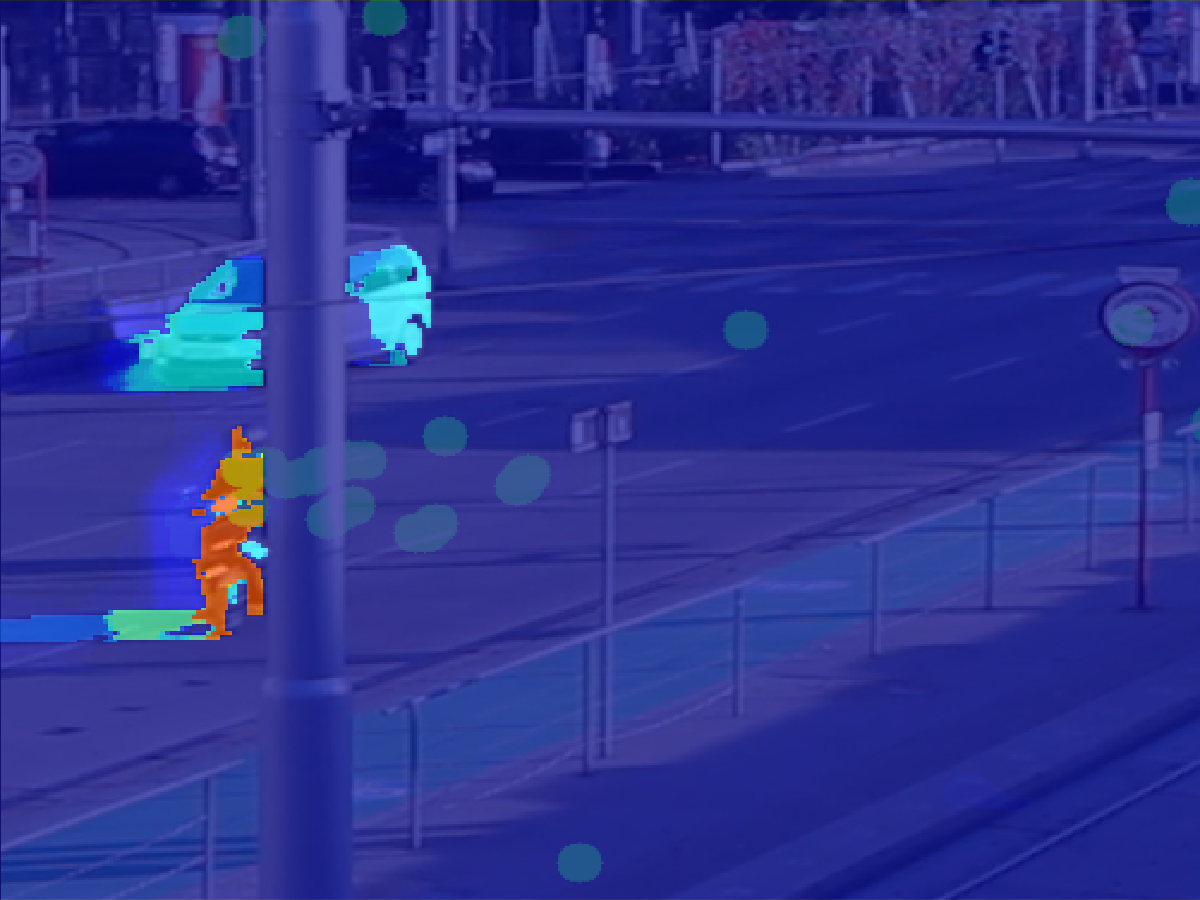}\\
\includegraphics[scale=0.196]{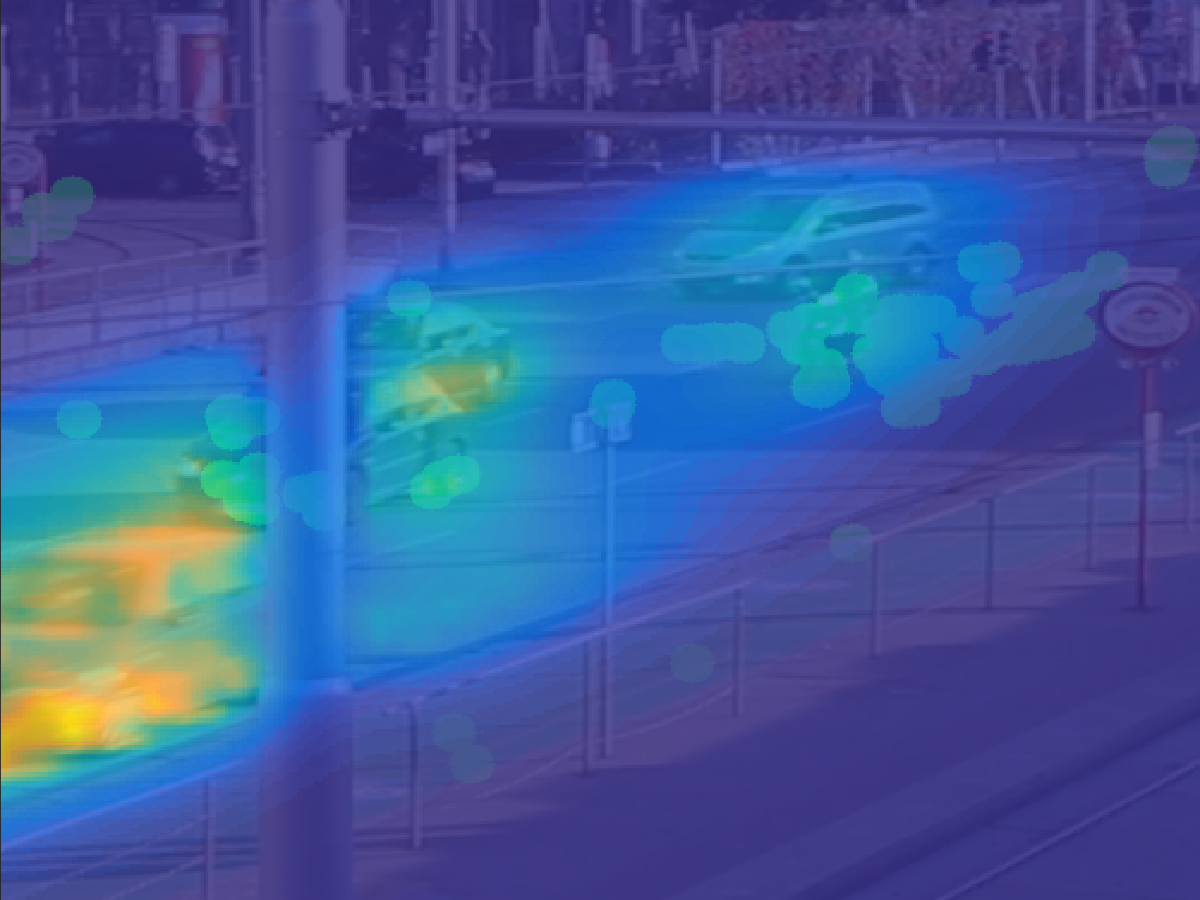} 
\includegraphics[scale=0.196]{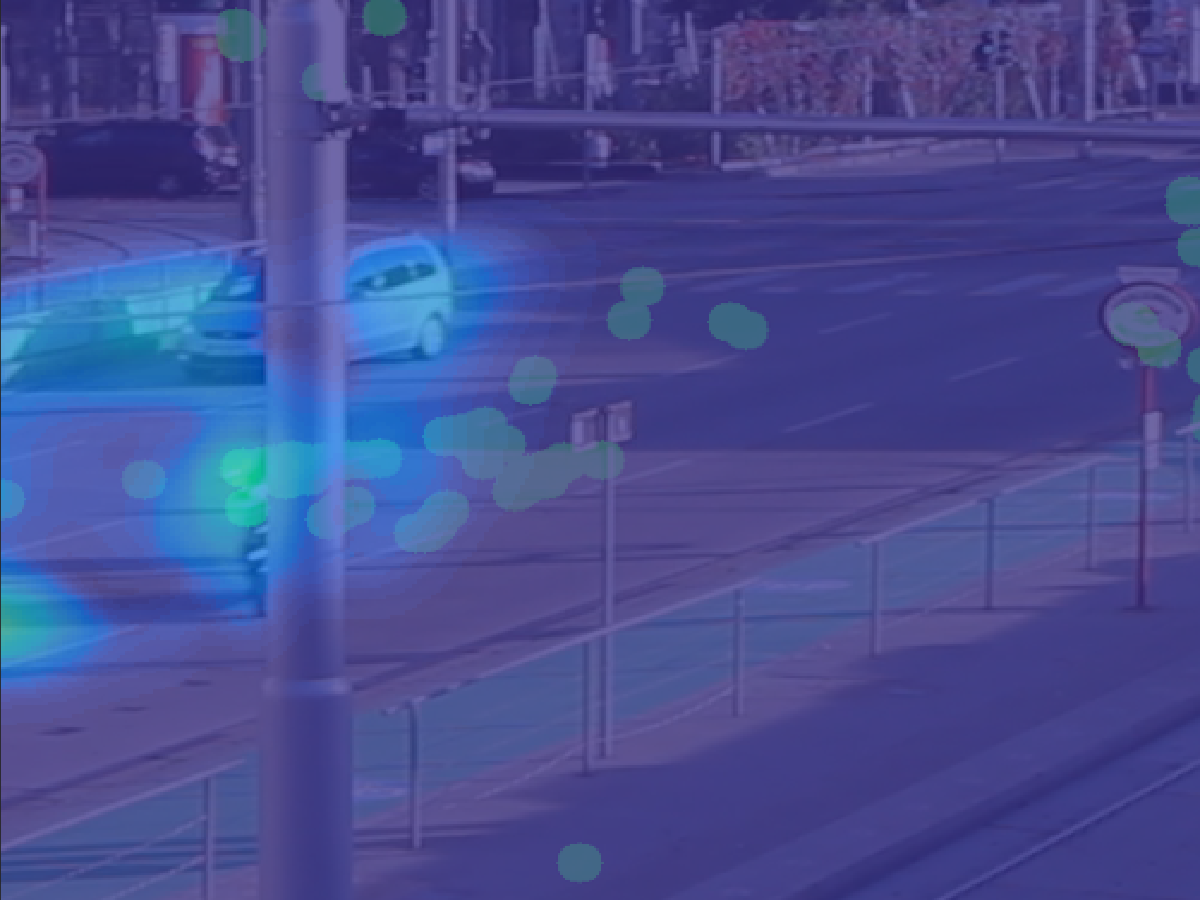} 
\includegraphics[scale=0.196]{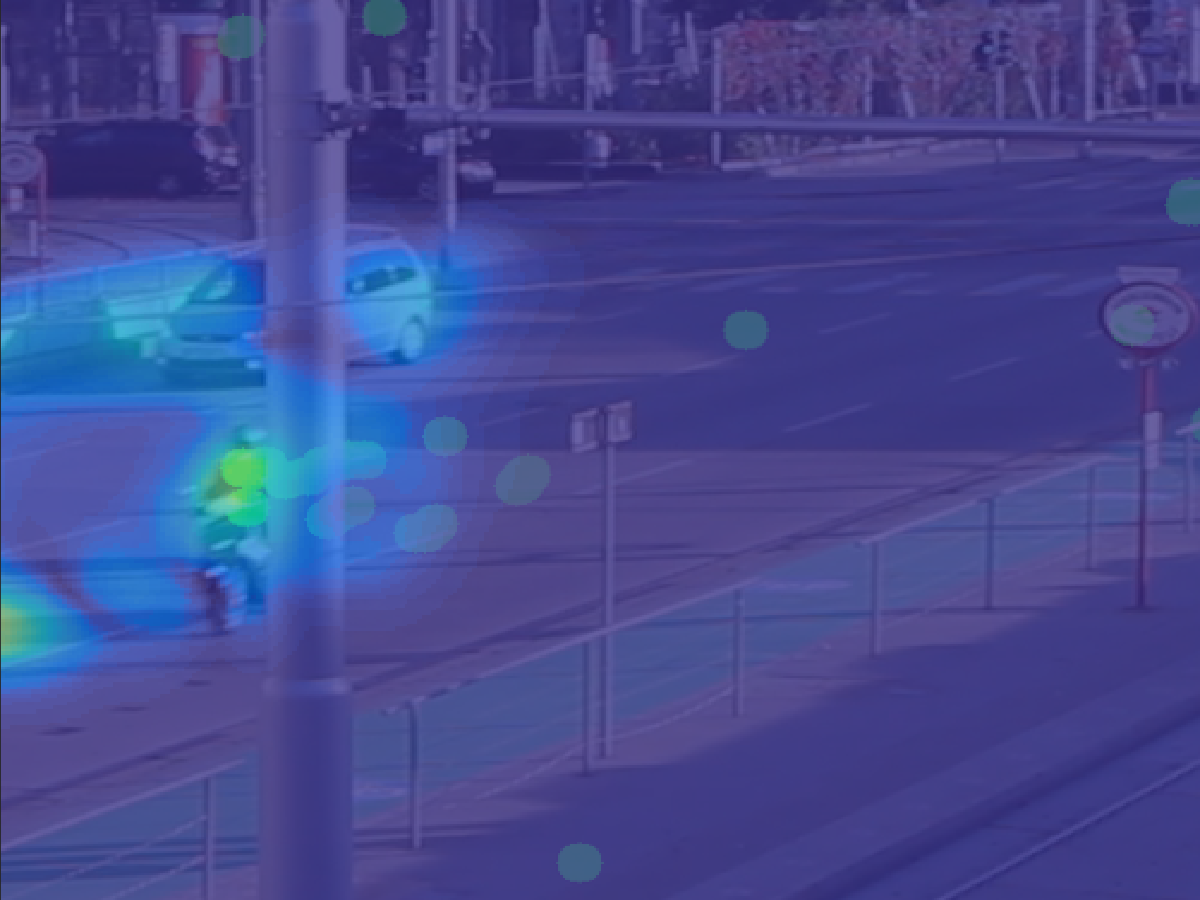} 
\caption{
The dynamic sequence shows traffic in a public street. A motorcycle is riding behind a pole and it is an object of interest. 
We notice that when the motorcycle is occluded (central column), only \eqref{eq:om} with saliency complemented data (top) is 
able to recognize the occluded part as salient. The method considering only two frames \protect\cite{SubRotBla14} (middle) 
does not recognize this area as salient. 
The method \eqref{eq:om} without complemented data (bottom) results in a lower correlation between the saliency map and the fixation distribution.}
\label{fig:exp5035}
\end{figure}
We compare our model \eqref{eq:om} with saliency complemented data, with a standard optical flow algorithm \cite{SubRotBla14} and with \eqref{eq:om} without complemented data. This last approach is shown to highlight the effect of complementing data with spatial saliency on the calculation of a dynamic saliency map.
In order to make a fair comparison between methods, for both our model \eqref{eq:om} with gray valued images complemented with saliency and the model of \cite{SubRotBla14} we set the amount of regularization (i.e. $\alpha$) to the value of forty.
We set $\alpha=30$ for \eqref{eq:om} using color valued images complemented with saliency. 
Finally, for \eqref{eq:om} with both types of complemented data, we set the factor enforcing smoothness over time (i.e $\lambda$) to the value of ten.
The parameters for \eqref{eq:om} without complemented data are like the one  for \eqref{eq:om} with complemented data.
In Figure \ref{fig:exp5035}, \ref{fig:exp5050}  and \ref{fig:exp5121} we present three sequences with occlusions. 
For each sequence we show the results of the models in one frame before, one during, and one after the occlusion. 
On every depicted frame, we superimposed partecipants' fixations, with green dots, of the last five frames before until the last five frames after the depicted frame.
For the proposed method \eqref{eq:om} with complemented data or without complemented data (see 
Figures \ref{fig:exp5035}, \ref{fig:exp5050} and \ref{fig:exp5121} [top and bottom]), the resulting saliency maps are similar for 
gray valued or color valued images. Therefore, we display only the gray valued version.

The first video in Figure \ref{fig:exp5035} shows a motorcycle riding behind a pole. We note that people 
are looking at the pole in order to follow the motion during the occlusion. A two-frame method such as \cite{SubRotBla14} 
does not recognize the pole as salient, while our method \eqref{eq:om} does so for the  occluding parts of the pole 
(see Figure \ref{fig:exp5035} [top]). Moreover, \eqref{eq:om} without complemented data recognizes the occlusion 
as salient, but the area is
not as strongly marked as attractive for attention 
compared to \eqref{eq:om} with complemented data. Indeed, 
we are not able to predict if the people are looking at the car or at the motorcycle in 
Figure \ref{fig:exp5035}. Therefore, the usage of complemented data in \eqref{eq:om} results in a better fit to the measured fixations 
compared to \eqref{eq:om} without complemented data (see Figure \ref{fig:exp5050} [top and bottom])
 We notice moreover that the method in \cite{SubRotBla14} discards the information regarding the motorcycle in the central 
 column of Figure \ref{fig:exp5035}.

In the second video in Figure \ref{fig:exp5050}, we see a woman running which is occluded by a tree for some moments. 
The woman is highly salient due to the strong color contrast of her red coat.
Her saliency is not recognized by \cite{SubRotBla14} while the woman is being occluded by a tree. 
Using both types of complemented data, our method \eqref{eq:om} recognizes her saliency correctly. We notice how a two-frame method like 
\cite{SubRotBla14}, in the central column of Figure \ref{fig:exp5050}, is affected by over-smoothing, 
which results in a wrong interpretation. This image illustrates that the temporal coherence inherent in \eqref{eq:om} but lacking in
\cite{SubRotBla14} makes the method more robust against over-smoothing. 
Also in this experiment like in the previous one, the usage of complemented data results in a saliency map more correlated to the 
gaze points compared to the one without complemented data (see Figure \ref{fig:exp5050} [top and bottom]).
 
Finally, the third video shows a couple walking behind a tree. This sequence is particularly challenging because it includes many 
points with light reflections and noise. 
We notice that this does not affect the proposed model \eqref{eq:om}, but it influences the result of a two-frame model like 
\cite{SubRotBla14}. 
This outcome shows that using temporal
coherence, as is done in our model \eqref{eq:om}, results in a robust method, usable for real video sequences.
The 
saliency of the walking couple is correctly recognized by our model \eqref{eq:om} using complemented data ( see Figure \ref{fig:exp5121} 
[top]). 
Moreover, the model marks the tree section as potentially attractive for
attention, a region that the participants indeed fixate during the occlusion (see Figure \ref{fig:exp5121} central column
[top]). 
We notice that in all these experiments the  proposed model \eqref{eq:om} with complemented data recognizes the salient part of the sequences more clearly compared to \eqref{eq:om} without complemented data and to \cite{SubRotBla14}.  

\subsubsection{Model evaluation measures}
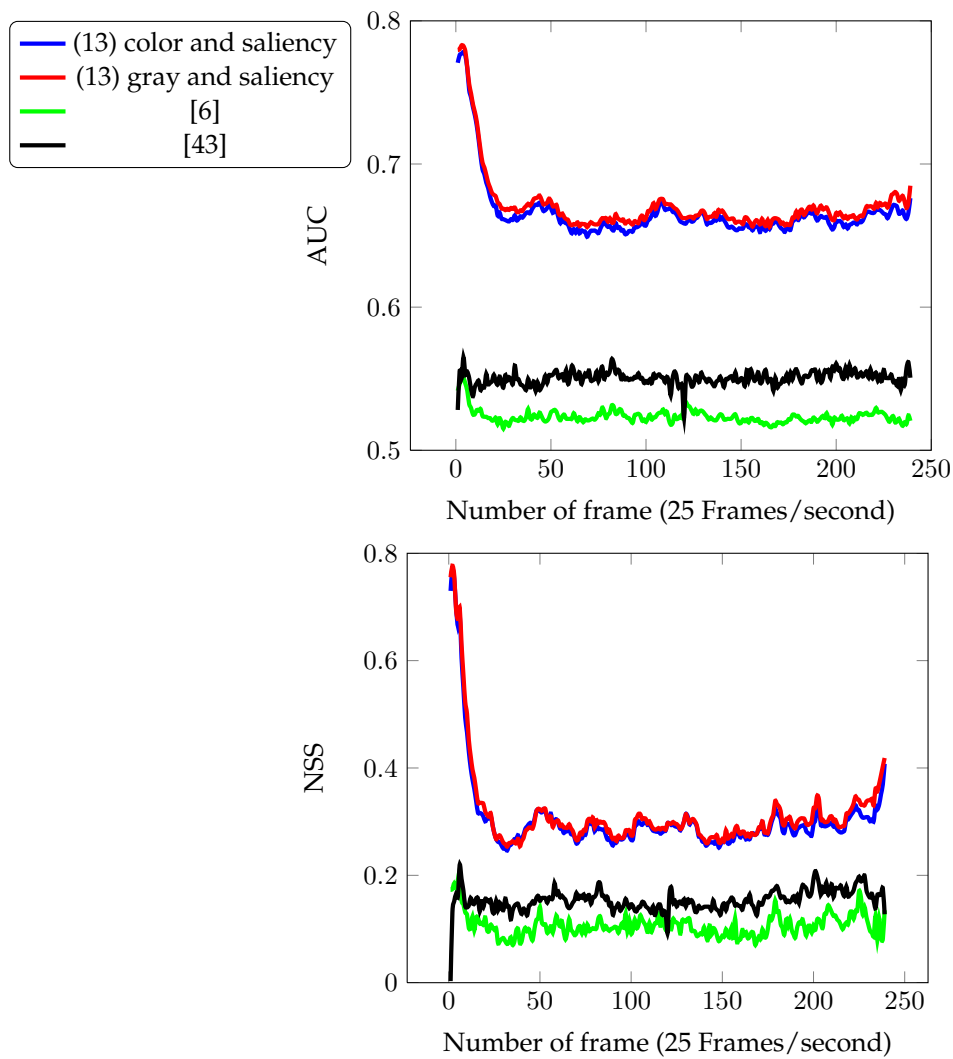
\begin{figure}[tp]
\centering
\hspace{-3.5cm}
\begin{tikzpicture}
\begin{axis}[xlabel={Number of frame (25 Frames/second)},ylabel={AUC},legend style={at={(axis cs:-55,0.8)},anchor=north east,rounded corners=3pt,thin},xmax=250,ymax=0.8,ymin=0.5]
\addplot[smooth,ultra thick,blue] table [x index=0, y index=1, col sep=comma] {AUC_Comparison_N};
\addlegendentry{\eqref{eq:om} color and saliency}
\addplot[smooth,ultra thick,red] table [x index=0, y index=2, col sep=comma] {AUC_Comparison_N};
\addlegendentry{\eqref{eq:om} gray and saliency}
\addplot[smooth,ultra thick,green] table [x index=0, y index=3, col sep=comma] {AUC_Comparison_N};
\addlegendentry{\cite{GuoZha10}}%
\addplot[smooth,ultra thick,black] table [x index=0, y index=4, col sep=comma] {AUC_Comparison_N};
\addlegendentry{\cite{SubRotBla14}}%
\end{axis}
\end{tikzpicture}
\begin{tikzpicture}
\begin{axis}[xlabel={Number of frame (25 Frames/second)},ylabel={NSS},ymin=0,ymax=0.8]
\addplot[smooth,ultra thick,blue] table [x index=0, y index=1, col sep=comma] {NSS_Comparison_N};
\addplot[smooth,ultra thick,red] table [x index=0, y index=2, col sep=comma] {NSS_Comparison_N};
\addplot[smooth,ultra thick,green] table [x index=0, y index=3, col sep=comma] {NSS_Comparison_N};
\addplot[smooth,ultra thick,black] table [x index=0, y index=4, col sep=comma] {NSS_Comparison_N};
\end{axis}
\end{tikzpicture}
\caption{Comparison according to AUC (top) and NSS (bottom) for the proposed approach \eqref{eq:om} with complemented data, the motion as modeled in \protect\cite{GuoZha10} and a standard optical flow algorithm \protect\cite{SubRotBla14}}
\label{fig:statMeas}
\end{figure}
As comparison metrics we chose \cite{RicDuvManGosDut13}: area under the curve (AUC) of the receiver operating characteristic (ROC) and normalized scanpath saliency (NSS).
The AUC treats the saliency map as a classifier.
Starting from a saliency map, the AUC algorithm labels all the pixels over a threshold as True Positive (TP) and all the pixels below as False Positive (FP). Human fixations are used as ground truth.
This procedure is repeated one hundred times with different threshold values. Finally, the algorithm can estimate the ROC curve and compute the AUC score. A perfect prediction corresponds to a score of 1.0 while a random classification results in 0.5.\\
The NSS was introduced by \cite{PetIyeeIttKoc05}.
For each point along a subject scan-path we extract the corresponding position $p$ and a partial value is calculated:
\begin{equation}
\label{eq:nssp}
NSS(p)=\frac{SM(p)- \mu_{SM}}{\sigma_{SM}}
\end{equation}
where $SM$ is the saliency map. 
In \eqref{eq:nssp} we normalize the saliency map in order to have zero mean and unit standard deviation.
The final NSS score is the average of $NSS(p)$:
\begin{equation}
NSS=\frac{1}{N}\sum_{p=1}^N NSS(p)
\end{equation}
where $N$ is the total number of fixation points.
The NSS, due to the initial normalization of the saliency map, allows comparison across different subjects.
For this measurement the perfect prediction corresponds to a score of 1.0.

In this paper, for the implementation of AUC and NSS we follow \cite{JudDurTor12}.\\
These measures are used to test the prediction performances of our \emph{new} dynamic saliency map. In other approaches \cite{MarHoGraGuyPelGue08,GuoZha10,IttBal05a,Oue04,PetItt07,ZhaTonCot09} the dynamic saliency maps are combined 
through a chosen weighting scheme \cite{MarHoGraGuyPelGue08,Oue04} with spatial saliency maps and then the resulting saliency map tested. Here, we do not discuss the choice of a weighting scheme and we test directly the dynamic saliency map.\\ 
In Figure \ref{fig:statMeas}, we compare the proposed model \eqref{eq:om}  using the two types of complemented data, 
the motion as modeled in \cite{GuoZha10} and a standard optical flow algorithm \cite{SubRotBla14}. 
We set the regularization parameter $\alpha=40$ for \cite{SubRotBla14}.
We notice in Figure \ref{fig:statMeas} that the proposed model \eqref{eq:om} performs similarly with both types of complemented data. Moreover, it performs   better than the other models considered. It is worth noticing that two frames models such as \cite{SubRotBla14} and \cite{GuoZha10} have bad performances. This is coherent with our previous results. The algorithms of \cite{SubRotBla14} and \cite{GuoZha10} discards more information than  \eqref{eq:om}  
(as the motorcycle in the central column of Figure  \ref{fig:exp5035}, second row). 
Moreover, the models of \cite{SubRotBla14} and of \cite{GuoZha10} fail in particular cases as the occlusion one described in the previous paragraph.

\section{Conclusion}
\label{sec:concl}
We have proposed a novel dynamic saliency map based on a variational approach and optical flow computations. 
The framework is applicable to every type of spatial saliency algorithm and results in significant improvements 
of model performance with regard to predicting human fixations in videos. We analyzed the possibility to use gray 
valued images or color valued images complemented with spatial saliency as input of our model. Finally, we studied 
the contribution of temporal coherence for calculating dynamic saliency maps and presented an application regarding occlusions. 
The results underline better performances (AUC and NSS) explaining visual attention compared to other approaches in literature 
\cite{GuoZha10,SubRotBla14}. 
\section{Acknowledgements}
This work is carried out within the project \emph{"Modeling Visual Attention as a Key Factor in Visual Recognition and Quality of Experience"} funded by the 
Wiener Wissenschafts und Technologie Fonds - WWTF (Grant no. CS11-009 to UA and OS).
OS is also supported by the Austrian Science Fund - FWF, Project S11704 with the national research network, 
NFN, Geometry and Simulation.
The authors would like to thank P. Elbau
for interesting discussions on optical flow. The authors also thank R. Seywerth for help with creating the stimulus videos and collecting the eye tracking data.

\appendix
\section{The use of complemented data results in a more reliable optical flow}
\label{app:aae}
We would like to emphasize that the purpose of our model \eqref{eq:om} is to estimate the 
dynamic saliency map of a movie sequence and not calculating a precise optical flow. 
As shown in Table \ref{tab:compCandG}, if we use saliency complemented data in \eqref{eq:om}, 
we can calculate the optical flow in imaging regions without regularization. In turn this means 
that by taking this into account we can reduce the amount of regularization (smaller $\alpha$) in 
optical flow minimization algorithms. 
\begin{figure}
\begin{center}
\includegraphics[scale=0.196]{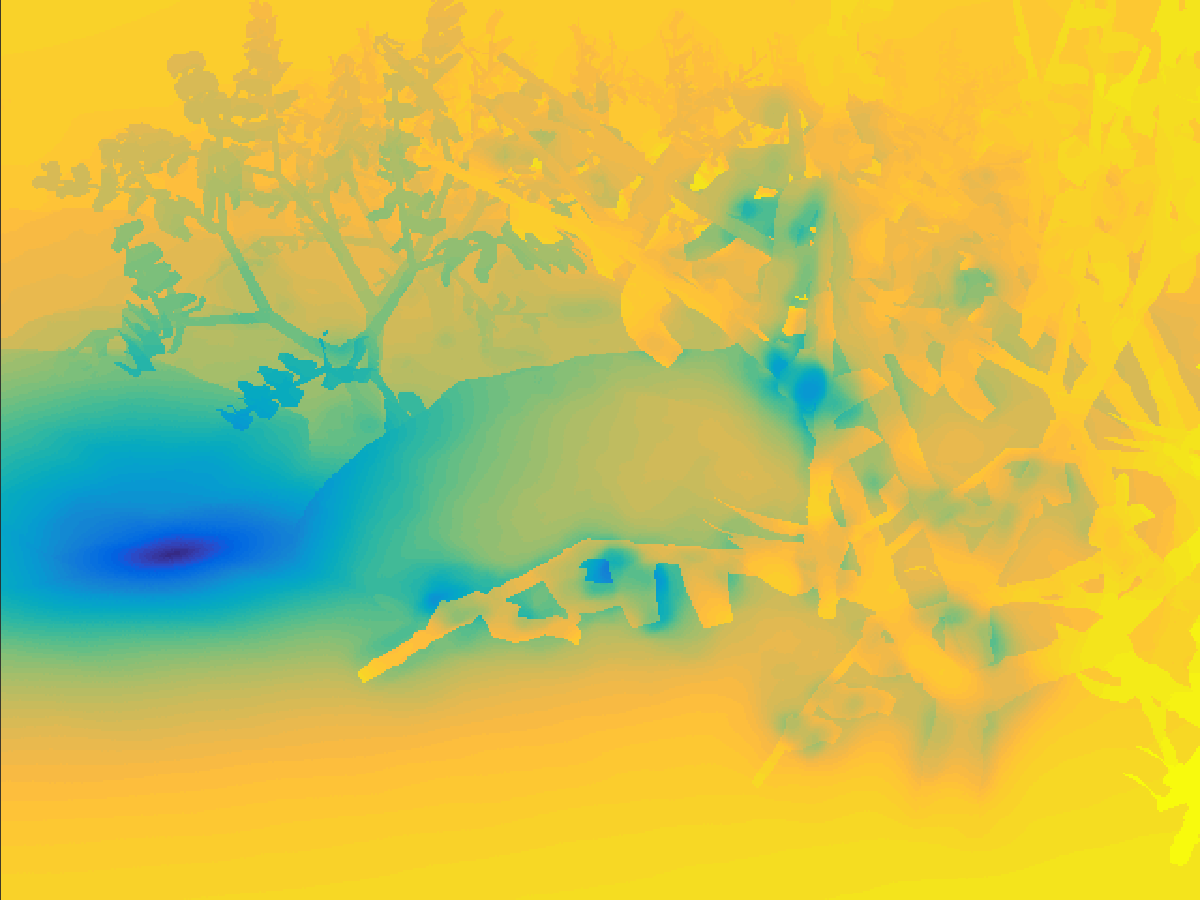}
\includegraphics[scale=0.196]{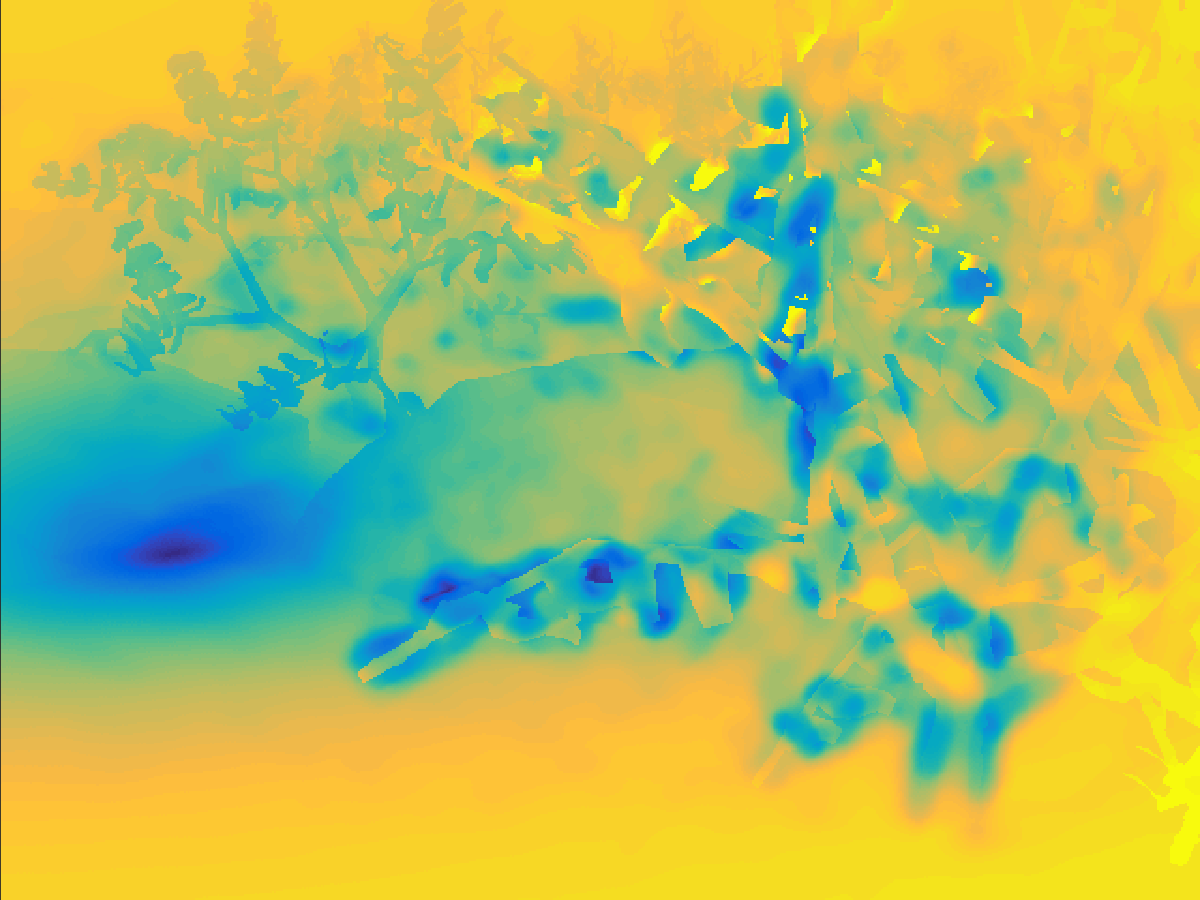}
\includegraphics[scale=0.196]{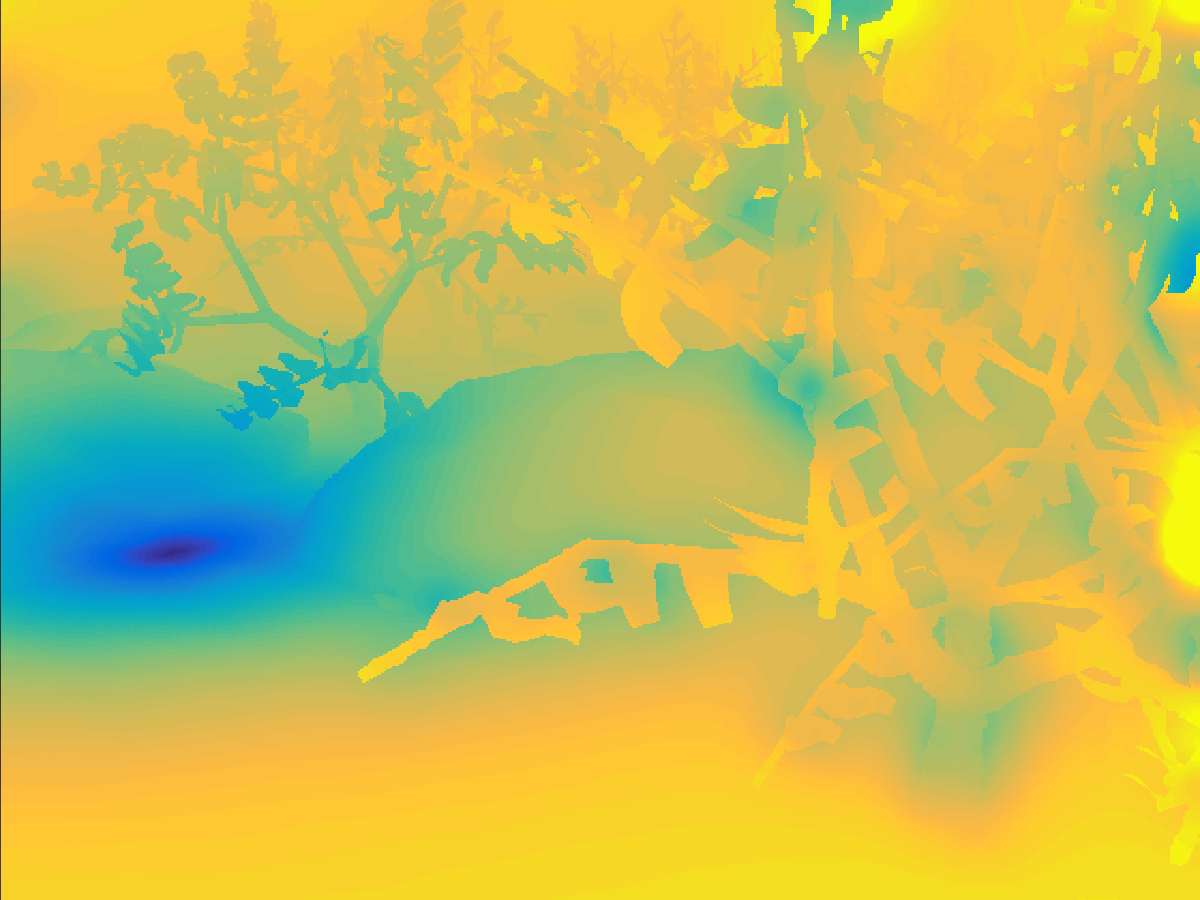}
\includegraphics[scale=0.196]{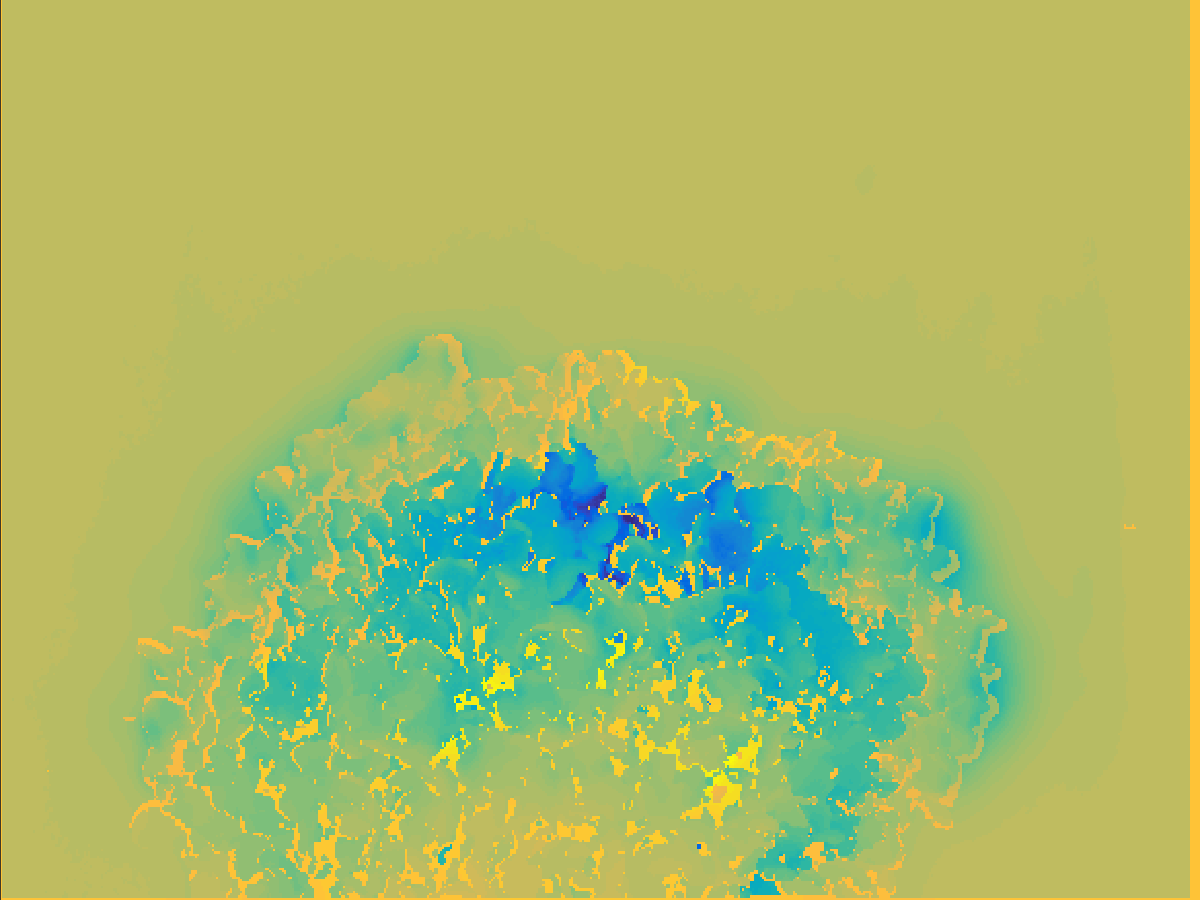} 
\includegraphics[scale=0.196]{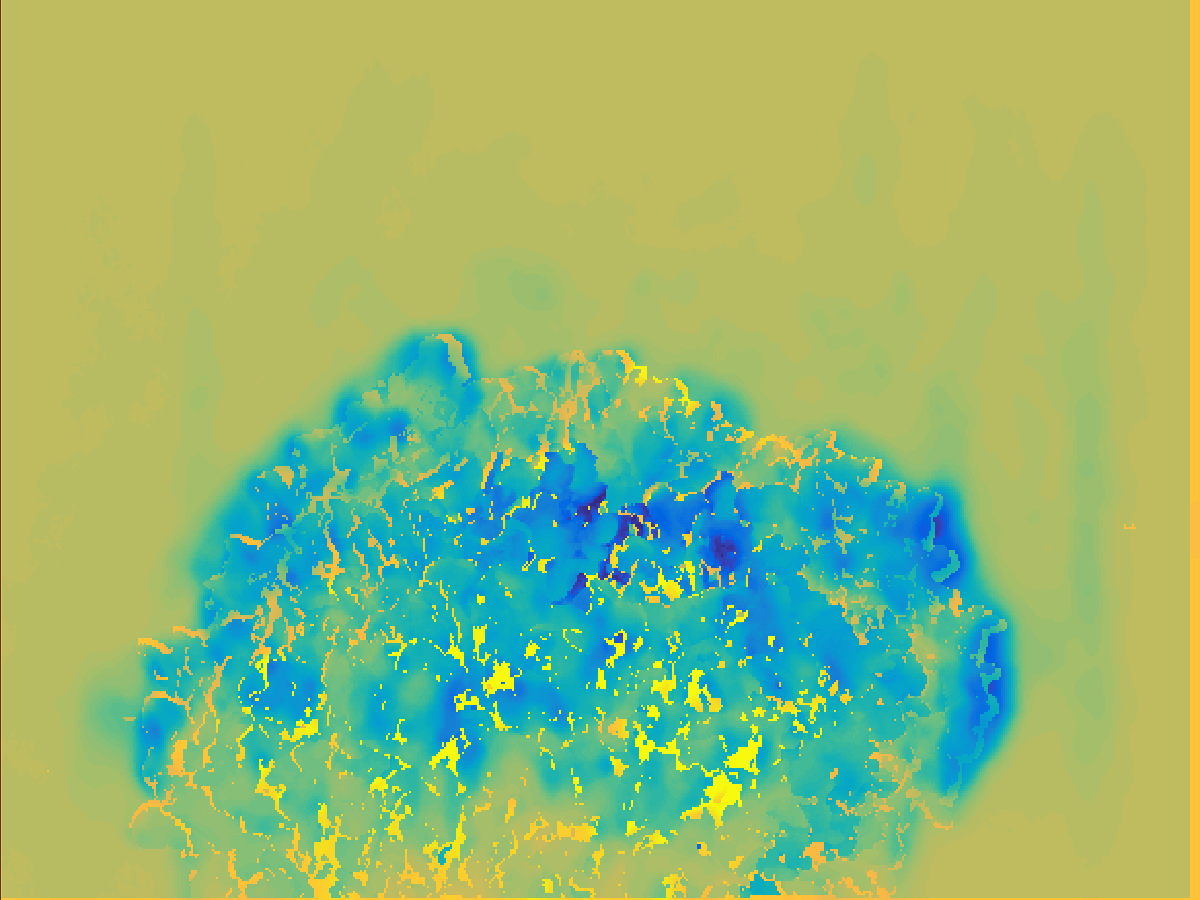}
\includegraphics[scale=0.196]{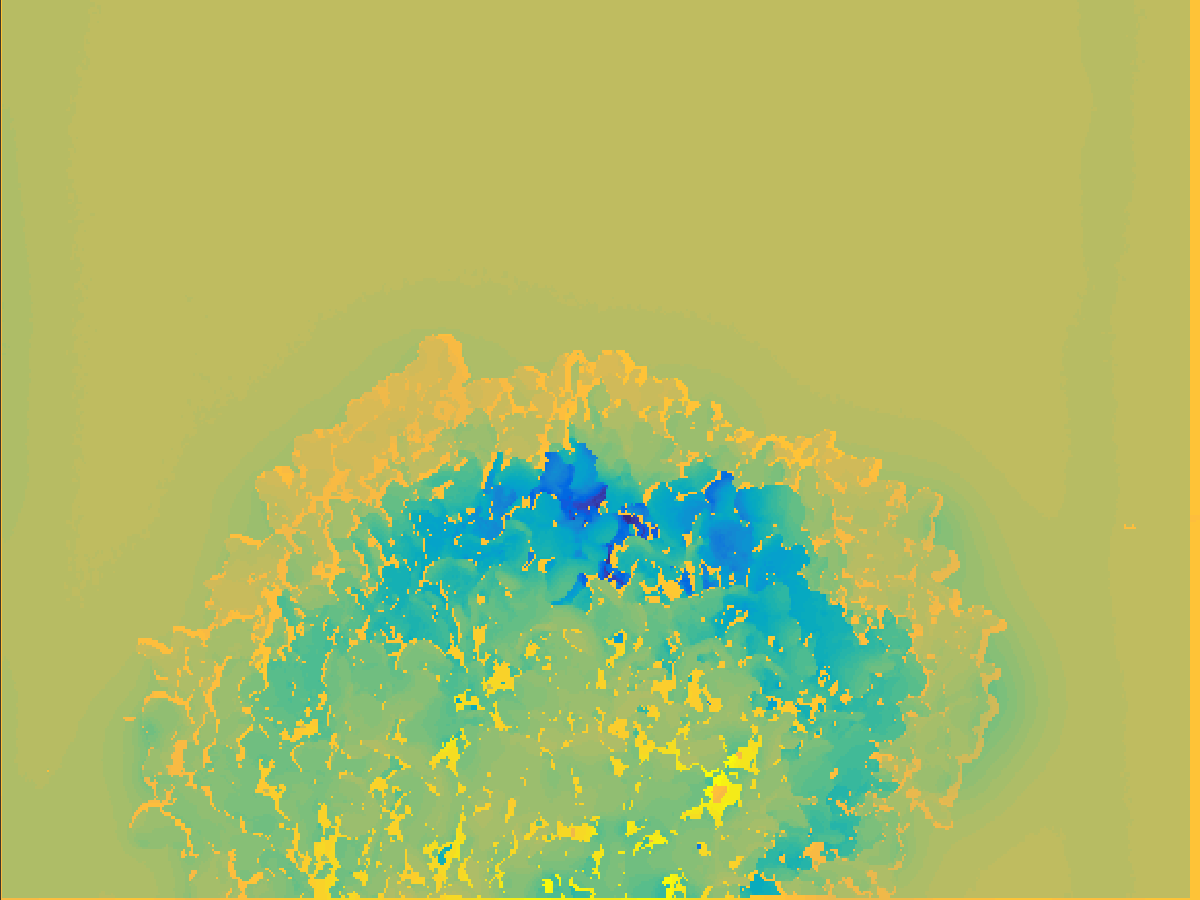} 
\end{center}
\caption{
We test the reliability of the optical flow comparing \eqref{eq:om} using gray valued images complemented with spatial saliency (left column), \eqref{eq:om} using color valued images complemented with spatial saliency (central column) and \eqref{eq:om} using color valued images not complemented (right column). The used comparison measure is the average angular error \protect\cite{BakSchRotBlaSze11} for the Grove 3 (top) and the Hydrangea (bottom) sequences from the Middleburry dataset. Blue shade color indicates points with small average angular error. 
For these experiments we set $\tau=0.01$ and $\alpha=0.8$, $\alpha=0.01$ and $\alpha=1.5$, respectively.}
\label{fig:caae}
\end{figure}
We test the assumptions above for two sequences of the Middlebury dataset \cite{BakSchRotBlaSze11}. We compare the 
flow obtained with our complemented data, with the true one, also called \emph{ground truth} for these sequences. 
We use the average angular error \cite{BarFleBea94} as comparison measure. 
In Figure \ref{fig:caae}, areas colored in shades of blue are the ones for which the average angular error is small. 
This means that in these areas, the flow is close to the true one.

In accordance with the results in Table \ref{tab:compCandG}, we set $\alpha$ to a value of: $0.8$ for gray scale data 
complemented with saliency, $0.01$ for color valued images complemented by saliency, and $1.5$ for color valued images. 
Ideally, we should obtain similar areas with small error.

We notice in Figure \ref{fig:caae} that the result we obtain by using color valued images complemented with saliency 
outperforms the other approaches. Moreover, the model \eqref{eq:om} that uses gray valued images complemented with 
saliency performs slightly performs better than the one using only color valued images without saliency. This exemplary 
test confirms our results in Table \ref{tab:compCandG}. 


\end{document}